\documentclass[runningheads,table,xcdraw]{llncs}
\usepackage{graphicx}
\usepackage{tikz}
\usepackage{comment}
\usepackage{amsmath,amssymb} 
\usepackage{color}
\colorlet{blue}{black}
\colorlet{red}{black}
\colorlet{purple}{black}

\usepackage[accsupp]{axessibility}  

\usepackage{pifont}
\usepackage{booktabs}
\usepackage{multirow}
\usepackage{ulem}
\usepackage{adjustbox}
\usepackage{float} 

\newcommand\blfootnote[1]{%
    \begingroup
    \renewcommand\thefootnote{}\footnote{#1}%
    \addtocounter{footnote}{-1}%
    \endgroup
}

\begin{document}
\pagestyle{headings}
\mainmatter
\def\ECCVSubNumber{3179}  

\title{Dense Cross-Query-and-Support Attention Weighted Mask Aggregation for Few-Shot Segmentation} 

\titlerunning{Dense Attention-Weighted Mask Aggregation for Few-Shot Segmentation}
%
\author{Xinyu Shi\inst{1}$^*$ \and
Dong Wei\inst{2}$^*$ \and
Yu Zhang\inst{1}$^\dagger$ \and
Donghuan Lu\inst{2}$^\dagger$ \and
Munan Ning\inst{2} \and
Jiashun Chen\inst{1} \and
Kai Ma\inst{2} \and
Yefeng Zheng\inst{2}}
\authorrunning{X. Shi et al.}
%
\institute{School of Computer Science \& Engineering, Key Lab of Computer Network \& Information Integration (Ministry of Education), Southeast Univ., Nanjing, China\\
\email{\{shixinyu,zhang\_yu,jiashunchen\}@seu.edu.cn}\\
Tencent Jarvis Lab, Shenzhen, China\\
\email{\{donwei,caleblu,masonning,kylekma,yefengzheng\}@tencent.com}}
\maketitle

\begin{abstract}
Research into Few-shot Semantic Segmentation (FSS) has attracted great attention, with the goal to segment target objects in a query image given only a few annotated support images of the target class.\blfootnote{$^*$Equal contributions and the work was done at Tencent Jarvis Lab.}
A key to this challenging task is to fully utilize the information in the support images by exploiting fine-grained correlations between the query and support images.\blfootnote{$^\dagger$Corresponding authors.}
However, most existing approaches either compressed the support information into a few class-wise prototypes, or used partial support information (e.g., only foreground) at the pixel level, causing non-negligible information loss.
In this paper, we propose Dense pixel-wise Cross-query-and-support Attention weighted Mask Aggregation (DCAMA), where both foreground and background support information are fully exploited via multi-level pixel-wise correlations between paired query and support features.
Implemented with the scaled dot-product attention in the Transformer architecture, DCAMA treats every query pixel as a token, computes its similarities with all support pixels, and predicts its segmentation label as an additive aggregation of all the support pixels’ labels---weighted by the similarities.
Based on the unique formulation of DCAMA, we further propose efficient and effective one-pass inference for \textit{n}-shot segmentation, where pixels of all support images are collected for the mask aggregation at once.
Experiments show that our DCAMA significantly advances the state of the art on standard FSS benchmarks of PASCAL-5$^i$, COCO-20$^i$, and FSS-1000, e.g., with 3.1\%, 9.7\%, and 3.6\% absolute improvements in 1-shot mIoU over previous best records.
Ablative studies also verify the design DCAMA.

\end{abstract}

\section{Introduction}
\label{sec:intro}
Recent years
Deep Neural Networks (DNNs) have achieved remarkable progress in semantic segmentation\cite{minaee2021image,taghanaki2021deep}, one of the fundamental tasks in computer vision.
However, the success of DNNs relies heavily on large-scale datasets, where abundant training images are available for every target class to segment.
In the extreme low-data regime, DNNs' performance may degrade quickly on previously unseen classes with only few examples due to poor generalization.
Humans, in contrast, are capable of learning new tasks rapidly in the low-data scenario, utilizing prior knowledge accumulated from life experience~\cite{lake2011Omniglot}.
Few-Shot Learning (FSL)~\cite{fei2006one,fink2005object} is a machine learning paradigm that aims at imitating such generalizing capability of human learners, where a model can quickly adapt for novel tasks given only a few examples.
Specifically, a \textit{support} set containing novel classes with limited samples is given for the model adaption, which is subsequently evaluated on a \textit{query} set containing samples of the same classes.
FSL has been actively explored in the field of computer vision, such as image classification~\cite{vinyals2016matching}, image retrieval~\cite{triantafillou2017few}, image captioning and visual question answering~\cite{dong2018fast}, and semantic segmentation~\cite{dong2018few,liu2020part,lu2021simpler,min2021hypercorrelation,nguyen2019feature,snell2017prototypical,sun2021boosting,tian2020prior,wang2020few,wang2019panet,yang2020prototype,zhang2021self,zhang2019canet,zhang2021few,zhang2020sg}.
In this paper, we tackle the problem of Few-shot Semantic Segmentation (FSS).

The key challenge of FSS is to fully exploit the information contained in the small support set.
Most previous works followed the notion of prototyping~\cite{snell2017prototypical},
where the information contained in the support images was abstracted into class-wise prototypes via class-wise average pooling~\cite{wang2019panet,zhang2019canet,zhang2020sg} or clustering~\cite{liu2020part,yang2020prototype}, against which the query features were matched for segmentation label prediction.
Being originally proposed for image classification tasks, however, the prototyping may result in great loss of the precious information contained in the already scarce samples when applied to FSS.
Given the dense nature of segmentation tasks, \cite{min2021hypercorrelation,wang2020few,zhang2019pyramid} recently proposed to explore pixel-wise correlations between the query features and the foreground support features, avoiding the information compression in prototyping.
However, these methods totally ignored the abundant information contained in the background regions of the support images.
Zhang et al.~\cite{zhang2021few} took into account background support features as well when computing the pixel-level correlations;
yet they only considered sparse correlations with uniformly sampled support pixels, causing potential information loss of another kind.
Thus, no previous work has fully investigated dense pixel-wise correlations of the query features with both of the foreground and background support features for FSS.

\begin{figure}[t]
  \centering
   \includegraphics[trim=33 153 35 72, clip, width=.92\linewidth]{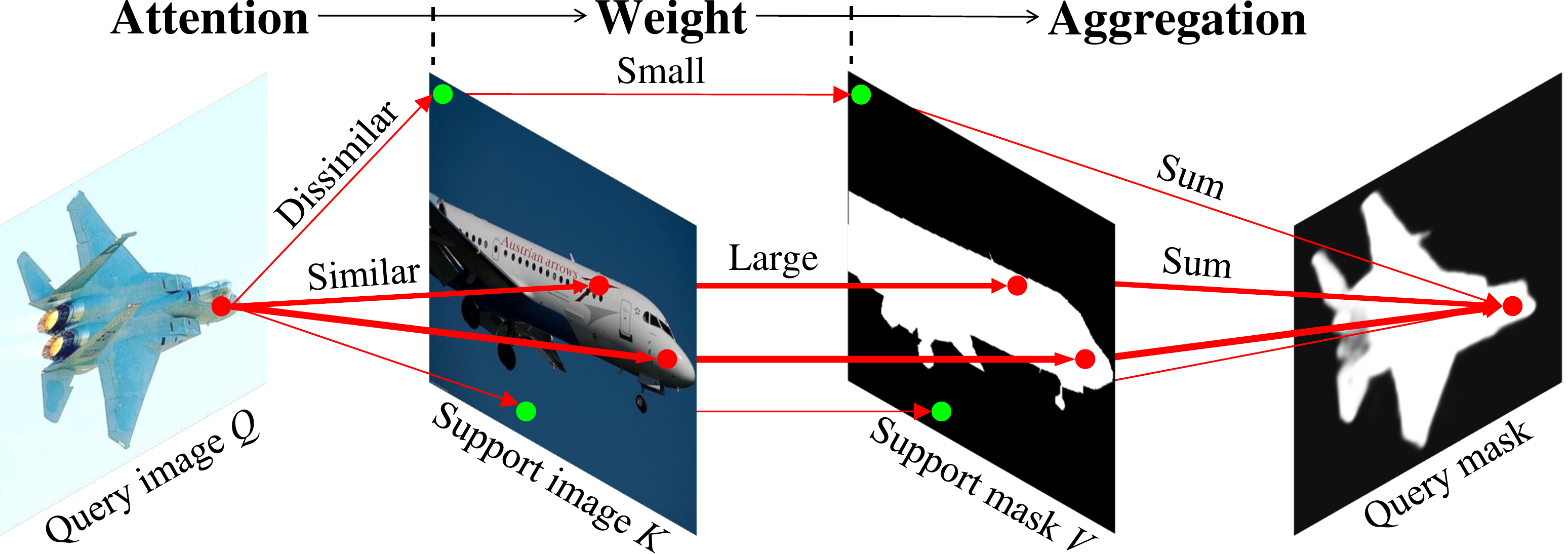}
   \caption{Conceptual overview of our approach.
   The query mask is directly predicted by a pixel-wise additive aggregation of the support mask values, weighted by the dense cross-query-and-support attention.}
   \label{fig:intro}
\end{figure}

In this work, we propose Dense pixel-wise Cross-query-and-support Attention weighted Mask Aggregation (DCAMA) for FSS, which fully exploits all available foreground and background features in the support images.
As shown in Fig. \ref{fig:intro}, we make a critical observation that the mask value of a query pixel can be predicted by an additive aggregation of the support mask values, proportionally weighted by its similarities to the corresponding support image pixels---including both foreground and background.
This is intuitive: if the query pixel is semantically close to a foreground support pixel, the latter will vote for foreground as the mask value for the former, and vice versa---an embodiment of metric learning~\cite{kulis2013metric}.
In addition, we notice that the DCAMA computation pipeline for all pixels of a query image can be readily implemented with the dot-product attention mechanism in the Transformer architecture~\cite{vaswani2017attention}, where each pixel is treated as a token, flattened features of all query pixels compose the query matrix $Q$,\footnote{Caution: not to confuse the query in FSL with that in Transformer.} those of all support image pixels compose the key matrix $K$, and flattened label values of the support mask pixels compose the value matrix $V$.
Then, the query mask can be readily and efficiently computed by $\operatorname{softmax}(QK^T)V$.
For practical implementation, we follow the common practices of multi-head attention~\cite{vaswani2017attention}, and multi-scale~\cite{lin2017feature} and multi-layer~\cite{min2021hypercorrelation} feature correlation;
{\color{purple}besides, the aggregated masks are mingled with skip-connected support and query features for refined query label prediction.}
As we will show, the proposed approach not only yields superior performance to that of the previous best performing method \cite{min2021hypercorrelation}, but also demonstrates higher efficiency in training.

Besides, previous works adopting the pipeline of pixel-level correlation paid little attention to extension from 1-shot to few-shot segmentation: they either performed separate 1-shot inferences followed by ensemble~\cite{min2021hypercorrelation},
or used a subset of uniformly sampled support pixels for inference~\cite{zhang2021few}.
Both solutions led to a loss of pixel-level information, due to the independent inferences prior to ensemble and drop of potentially useful pixels, respectively.
In contrast, we make full use of the support set, by using all pixels of all the support images and masks to compose the $K$ and $V$ matrices, respectively.
Meanwhile, we use the same 1-shot trained model for testing in different few-shot settings.
This is not only computationally economic but also reasonable, because what the model actually learns from training is a metric space for cross-query-and-support attention.
As long as the metric space is well learned, extending from 1-shot to few-shot is simply aggregating the query mask from more support pixels.

In summary, we make the following contributions:
\begin{itemize}
\item We innovatively model the FSS problem as a novel paradigm of Dense pixel-wise Cross-query-and-support Attention weighted Mask Aggregation (DCAMA), which fully exploits the foreground and background support information.
    Being essentially an embodiment of metric learning, the paradigm is nonparametric {\color{purple}in mask aggregation} and expected to generalize well.
\item We implement the new FSS paradigm with the well-developed dot-product attention mechanism in Transformer, for simplicity and efficiency.
\item Based on DCAMA, we propose an approach to $n$-shot inference that not only fully utilizes the available support images at the pixel level, but also is computationally economic without the need for training $n$-specific models for different few-shot settings.
\end{itemize}
The comparative experimental results on three standard FSS benchmarks of PASCAL-5$^i$~\cite{shaban2017one}, COCO-20$^i$~\cite{nguyen2019feature}, and FSS-1000~\cite{li2020fss} show that our DCAMA sets new state of the art on all the three benchmarks and in both 1- and 5-shot settings.
In addition, we conduct thorough ablative experiments to validate the design of DCAMA.

\section{Related Work}
\label{sec:rela}

\textbf{Semantic segmentation.}
Semantic segmentation is a fundamental task in computer vision, whose goal is to classify each pixel of an image into one of the predefined object categories.
In the past decade, impressive progress has been made with the advances in DNNs~\cite{minaee2021image,taghanaki2021deep}.
The cornerstone Fully Convolutional Network (FCN)~\cite{long2015fully} proposed to replace the fully connected output layers in classification networks with convolutional layers to efficiently output pixel-level dense prediction for semantic segmentation.
Since then, prevailing segmentation DNN models~\cite{chen2017deeplab,ronneberger2015u,zhao2017pyramid} have evolved to be dominated by FCNs with the generic encoder-decoder architecture, where
techniques like skip connections ~\cite{ronneberger2015u} and multi-scale processing~\cite{chen2018encoder,zhao2017pyramid} are commonly employed for better performance.
Recently, inspired by the success of the Vision Transformer (ViT)~\cite{dosovitskiy2020image}, we have witnessed a surge of active attempts to apply the Transformer architecture to semantic segmentation~\cite{strudel2021segmenter,zhu2021unified}.
Notably, the Swin Transformer, a general-purpose computer-vision backbone featuring a hierarchical architecture and shifted windows, achieved new state-of-the-art (SOTA) performance on the ADE20K semantic segmentation benchmark~\cite{liu2021swin}. 
Although these methods demonstrated their competency given abundant training data and enlightened our work, they all suffered from the inability to generalize for the low-data regime.

\textbf{Few-shot learning.}
Few-shot learning~\cite{fei2006one} is a paradigm that aims to improve the generalization ability of machine learning models in the low-data regime.
Motivated by the pioneering work ProtoNet~\cite{snell2017prototypical}, most previous works~\cite{dong2018few,liu2020part,wang2019panet,yang2020prototype,zhang2019canet,zhang2020sg} on FSS followed a metric learning~\cite{dong2018few} pipeline, where the information contained in support images was compressed into abstract prototypes, and the query images were classified based on certain distances from the prototypes in the metric space.
Dong and Xing~\cite{dong2018few} extended the notion of prototype for FSS by computing class-wise prototypes via global average pooling on features of masked support images.
Instead of masking the input images, PANet~\cite{wang2019panet} performed average pooling on masked support features for prototyping and introduced prototype alignment as regularization.
CANet~\cite{zhang2019canet} also relied on masked feature pooling but followed the Relation Network~\cite{sung2018learning} to learn a deep metric using DNNs.
PFENet~\cite{tian2020prior} further proposed a training-free prior mask as well as a multi-scale feature enrichment module.
Realizing the limited representation power of a single prototype, \cite{cui2021unified,liu2020part,yang2020prototype,zhang2021self} all proposed to represent a class with multiple prototypes.
These prototype-based methods jointly advanced the research into FSS;
however, compressing all available information in a support image into merely one or few concentrated prototypes unavoidably resulted in great information loss.

More recently, researchers started to exploit pixel-level information for FSS, to better utilize the support information and align with the dense nature of the task.
PGNet~\cite{zhang2019pyramid} and DAN~\cite{wang2020few} modeled pixel-to-pixel dense connections between the query and support images with the graph attention~\cite{velivckovic2018graph}, whereas HSNet~\cite{min2021hypercorrelation} constructed 4D correlation tensors
to represent dense correspondences between the query and support images.
Notably, HSNet proposed center-pivot 4D convolutions
for efficient high-dimensional convolutions, and achieved SOTA performance on three public FSS benchmarks by large margins.
However, these methods all masked out background regions in the support images,
ignoring the rich information thereby.
In contrast, our DCAMA makes equal use of both foreground and background information.
In addition, implementing straightforward metric learning with the multi-head attention~\cite{vaswani2017attention}, our DCAMA is easier to train than HSNet, converging to higher performance in much fewer epochs and much less time.
Lastly, instead of the ensemble of separate 1-shot inferences~\cite{min2021hypercorrelation} or training $n$-specific models~\cite{zhang2019canet} for $n$-shot inference, DCAMA constructs the key and value matrices with pixels of all the support images and masks, and infers only once reusing the 1-shot trained model.

\textbf{Vision Transformers for FSS.}
Inspired by the recent success of the Transformer architecture in computer vision~\cite{dosovitskiy2020image,liu2021swin}, researchers also started to explore their applications to FSS lately.
Sun et al.~\cite{sun2021boosting} proposed to employ standard multi-head self-attention Transformer blocks
for global enhancement.
Lu et al.~\cite{lu2021simpler} designed the Classifier Weight Transformer (CWT) to dynamically adapt
the classifier's weights for each query image.
However, both of them still followed the prototyping pipeline thus did not fully utilize fine-grained support information.
The Cycle-Consistent Transformer (CyCTR)~\cite{zhang2021few} may be the most related work to ours in terms of: (i) pixel-level cross-query-and-support similarity computation with the dot-product attention mechanism, and (ii) use of both foreground and background support information.
The main difference lies in that CyCTR used the similarity to guide the reconstruction of query features from support features, which were then classified into query labels by a conventional FCN.
In contrast, our DCAMA {\color{purple}can} directly predict the query labels by aggregating the support labels weighted by this similarity, which is metric learning and expected to generalize well for its nonparametric form.
Another distinction is that CyCTR subsampled the support pixels
thus was subject to potential information loss dependent on the sampling rate, whereas our DCAMA makes full use of all available support pixels.

\begin{figure}[t]
  \centering
   \includegraphics[trim=0 8 0 5, clip, width=\textwidth]{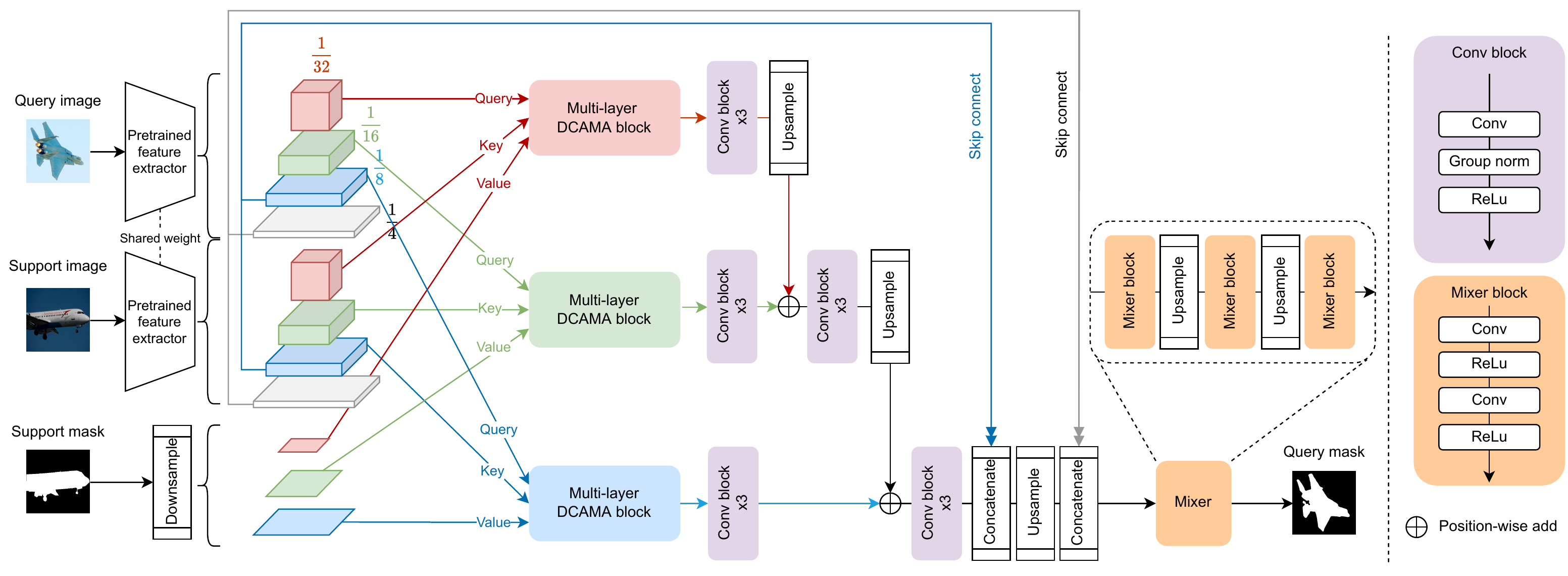}
   \caption{Pipeline of the proposed framework, shown in 1-shot setting.
   DCAMA: Dense Cross-query-and-support Attention weighted Mask Aggregation.}
   \label{fig:overview}
\end{figure}

\section{Methodology}
\label{sec:meth}

In this section, we first introduce the problem setup of Few-shot Semantic Segmentation (FSS).
Then, we describe our Dense Cross-query-and-support Attention weighted Mask Aggregation (DCAMA) framework in 1-shot setting.
Lastly, we extend the framework for $n$-shot inference.

\subsection{Problem Setup}
\label{subsec:problem setting}
In a formal definition, a 1-way $n$-shot FSS task $\mathcal{T}$ comprises a support set $\mathcal{S}=\{(I^s, M^s)\}$, where $I^s$ and $M^s$ are a support image and its ground truth mask, respectively, and $|S|=n$;
and similarly a query set $\mathcal{Q}=\{(I^q, M^q)\}$, where $\mathcal{S}$ and $\mathcal{Q}$ are sampled from the same class.
The goal is to learn a model to predict $M^q$ for each $I^q$ given the support set $\mathcal{S}$, subject to $n$ being small for few shots.
For method development, suppose we have two image sets $\mathcal{D}_\mathrm{train}$ and $\mathcal{D}_\mathrm{test}$ for model training and evaluation, respectively, where $\mathcal{D}_\mathrm{train}$ and $\mathcal{D}_\mathrm{test}$ do not overlap in classes.
We adopt the widely used meta-learning paradigm called episodic training~\cite{vinyals2016matching}, where each episode is designed to mimic the target task by subsampling the classes and images in the training set.
Specifically, we repeatedly sample new episodic tasks $\mathcal{T}$ from $\mathcal{D}_\mathrm{train}$ for model training.
The use of episodes is expected to make the training process more faithful to the test environment and thereby improve generalization~\cite{ravi2016optimization}.
For testing,
the trained model is also evaluated with episodic tasks but sampled from $\mathcal{D}_\mathrm{test}$.

\subsection{DCAMA Framework for 1-Shot Learning}

\textbf{Overview.}\label{subsec:overview}
The overview of our DCAMA framework is shown in Fig. \ref{fig:overview}.
For simplicity, we first describe our framework for 1-shot learning.
The input to the framework is the query image, and the support image and mask.
First, both the query and support images are processed by a pretrained feature extractor, yielding multi-scale query and support features.
Meanwhile, the support mask is downsampled to multiple scales matching the image features.
Second, the query features, support features, and support mask at each scale are input to the multi-layer DCAMA block of the same scale as $Q$, $K$, and $V$ for multi-head attention~\cite{vaswani2017attention} and aggregation of the query mask.
The query masks aggregated at multiple scales are processed and combined with convolutions, upsampling (if needed), and element-wise additions. 
Third, the output of the previous stage (multi-scale DCAMA) is concatenated with multi-scale image features via skip connections, and subsequently mingled by a mixer to produce the final query mask.
In the following, we describe each of these three stages in turn, with an emphasis on the second---which is our main contribution.

\textbf{Feature extraction and mask preparation.}
First of all, both the query and support images are input to a pretrained feature extractor
to obtain the collections of their multi-scale multi-layer feature maps $\{F^q_{i,l}\}$ and $\{F^s_{i,l}\}$, where $i$ is the scale of the feature maps with respect to the input images and $i\in\{\frac{1}{4}, \frac{1}{8}, \frac{1}{16}, \frac{1}{32}\}$ for the feature extractor we use, and $l\in\{1,\ldots,L_i\}$ is the index of all layers of a specific scale $i$.
Unlike most previous works that only used the last-layer feature map of each scale, i.e., $F_{i,L_i}$, we follow Min et al.~\cite{min2021hypercorrelation} to make full use of all intermediate-layer features, too.
Meanwhile, support masks of different scales $\{M^s_i\}$
are generated via bilinear interpolation from the original support mask.
The query features, support features, and support masks of scales $i\in \{\frac{1}{8}, \frac{1}{16}, \frac{1}{32}\}$ are input to the multi-layer DCAMA block,\footnote{The $\frac{1}{4}$ scale features are not cross-attended due to hardware constraint.}
as described next.

\begin{figure}[t]
  \centering
   \includegraphics[trim=0 2 0 0, clip, width=\textwidth]{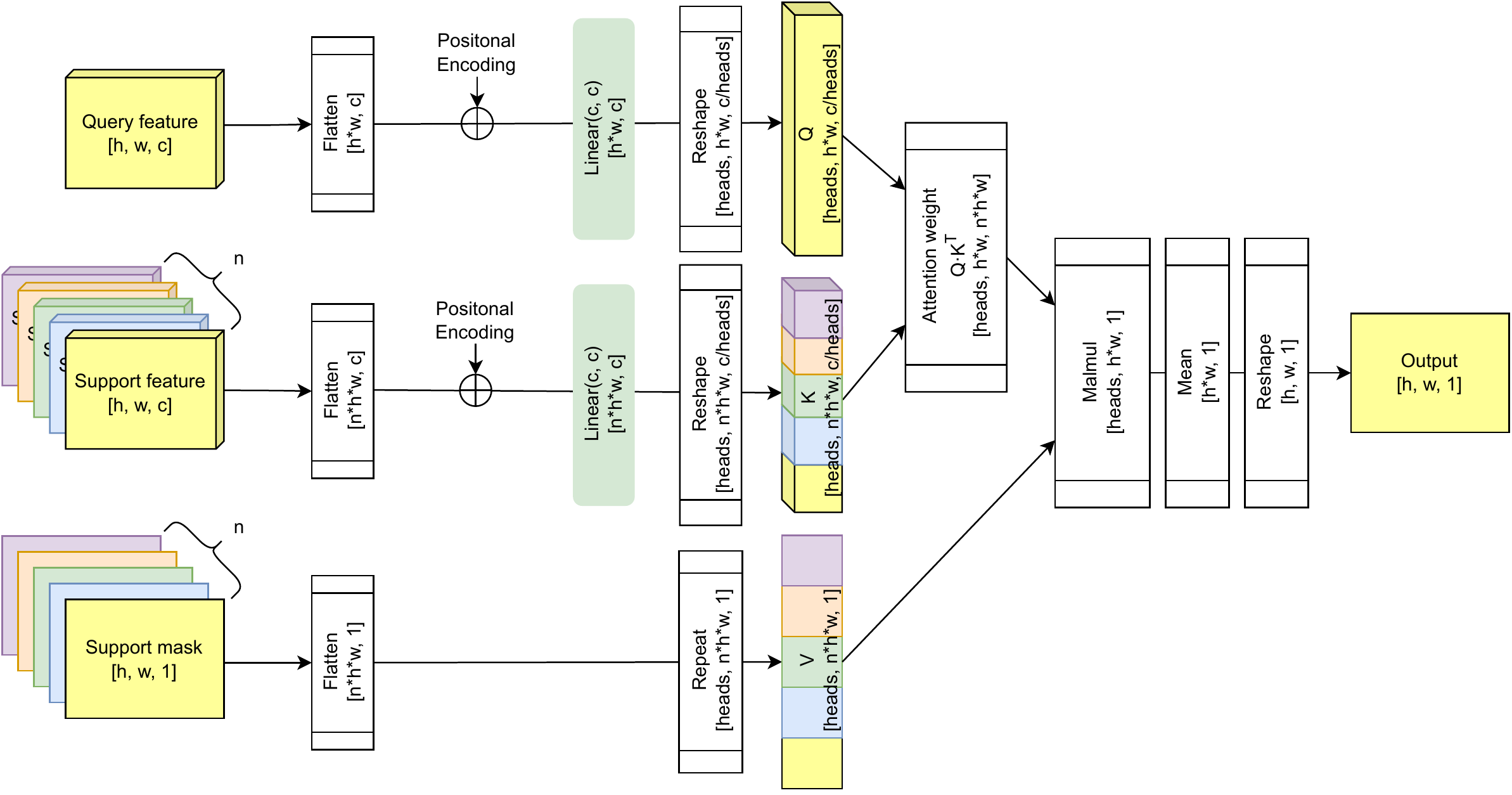}
   \caption{Dense Cross-query-and-support Attention weighted Mask Aggregation (DCAMA) for generic $n$-shot settings ($n\geq1$).}\label{fig:nshot}
\end{figure}

\textbf{Multi-scale multi-layer cross attention weighted mask aggregation.}\label{subsec:mha}
The scaled dot-product attention is the core of the Transformer~\cite{vaswani2017attention} architecture, and is formulated as:\vspace{-2.75mm}
\begin{equation}\vspace{-1.mm}
  \text{\phantom{WWW}}\operatorname{Attn}(Q, K, V)=\operatorname{softmax}\big({QK^T}\big/{\sqrt{d}}\big)V,
  \label{eq:attn}
\end{equation}
where $Q, K, V$ are sets of query, key, and value vectors packed into matrices, and $d$ is the dimension of the query and key vectors.
In this work, we adopt Eqn.~(\ref{eq:attn}) to compute dense pixel-wise attention across the query and support features, and subsequently weigh the query mask aggregation process from the support mask with the attention values.
Without loss of generality, we describe the mechanism with a pair of generic query and support feature maps $F^q, F^s\in\mathbb{R}^{h\times w\times c}$, where $h$, $w$, and $c$ are the height, width, and channel number, respectively, and a generic support mask $M^s\in\mathbb{R}^{h\times w\times 1}$ of the same size.
As shown in Fig. \ref{fig:nshot},
we first flatten the two-dimensional (2D) inputs
to treat each pixel as a token, and then generate the $Q$ and $K$ matrices from the flattened $F^q$ and $F^s$ after adding positional encoding and linear projections.
We follow the original Transformer~\cite{vaswani2017attention} to use sine and cosine functions of different frequencies for positional encoding, and employ the multi-head attention.
As to the support mask, it only needs flattening to construct $V$.
After that, the standard scaled dot-product attention in Eqn.~(\ref{eq:attn}) can be readily computed for each head.
Lastly, we average the outputs of the multiple heads for each token, and reshape the tensor to 2D to obtain $\hat{M}^q\in\mathbb{R}^{h\times w\times 1}$, which is the aggregated query mask.

\textit{Remarks.}
It is worth explaining the physical meaning of the DCAMA process.
For a specific query pixel, $QK^T$ measures its similarities to all the support pixels, and the subsequent multiplication with $V$ aggregates its mask value from the support mask, weighted by the similarities.
Intuitively, if it is more similar (closer) to the foreground than background pixels, the weighted aggregation process will vote for foreground for the pixel, and vice versa.
In this way, our DCAMA utilizes all support pixels---both foreground and background---for effective metric learning.

In practical implementation, we conduct DCAMA separately on query-support feature pairs $(F^s_{i,l}, F^q_{i,l})$ for all the intermediate and last layers of a specific scale $i$, and concatenate the set of independently aggregated query masks to get $\hat{M}^q_i=\operatorname{concat}\{\hat{M}^q_{i,l}|l=1,\ldots,L_i\}$.
The DCAMA operations on all-layer features of a specific scale followed by the concatenation compose a multi-layer DCAMA block (see Fig. \ref{fig:overview}), and we have three such blocks for scales $i\in\{\frac{1}{8}, \frac{1}{16}, \frac{1}{32}\}$, respectively.
Then, $\hat{M}^q_i$ is processed by three Conv blocks which gradually increase its channel number from $L_i$ to 128, upsampled by bilinear interpolation, combined with the counterpart of the one time larger scale by element-wise addition, and again processed by another three Conv blocks of a constant channel number.
The first three Conv blocks prepare $\hat{M}^q_i$ for effective inter-scale integration with that of the larger scale by the second three Conv blocks.
This process repeats from $i=\frac{1}{32}$ up to $\frac{1}{8}$, yielding a collection of intermediate query masks to be fused with skip-connected image features for final prediction.

\textbf{Mask-feature mixer.}
Motivated by the success of the skip connection design in generic semantic segmentation~\cite{ronneberger2015u,zhao2017pyramid}, we also propose to skip connect the image features to the output (upsampled when needed) of the previous stage via concatenation (Fig. \ref{fig:overview}).
Specifically, we skip connect the last-layer features at the $\frac{1}{4}$ and $\frac{1}{8}$ scales based on our empirical experiments (included in the supplementary material).
Then, the concatenated intermediate query masks and image features are fused by three mask-feature mixer blocks, each containing two series of convolution and ReLU operations.
The mixer blocks gradually decrease the number of output channels to 2 (for foreground and background, respectively) for 1-way segmentation, with two interleaved upsampling operations to restore the output size to that of the input images.

\subsection{Extension to $n$-Shot Inference}
So far we have introduced our DCAMA framework for 1-shot segmentation, next we extend it for $n$-shot setting, too.
Although it is possible to develop and train a specific model for each different value of $n$ (e.g., \cite{zhang2019canet}), it is computationally prohibitive to do so.
In contrast, many previous works extended the 1-shot trained model for $n$-shot inference without retraining~\cite{min2021hypercorrelation,tian2020prior,zhang2021self}.
The most common method is to perform $n$ 1-shot inferences separately with each of the support images, followed by certain ensembles of the individual inferences~\cite{min2021hypercorrelation,zhang2021self}.
However, this method unavoidably lost the pixel-level subtle cues across support images, as it treated each support image independently for inference.
In this work, we also reuse the 1-shot trained model for $n$-shot inference for computational efficiency, but meanwhile utilize all the pixel-wise information of all support images simultaneously during inference.

Thanks to the problem formulation of DCAMA, the extension is straightforward.
First, we obtain the multi-scale image features and masks for all the support images.
Next, we simply treat the additional pixels in the extra support features and masks as more tokens in $K$ and $V$, with a proper reshaping of the tensors (Fig. \ref{fig:nshot}).
Then, the entire DCAMA process (cross attention, mask aggregation, etc.) stays the same as the 1-shot setting.
This is feasible as the core of DCAMA is the scaled dot-product attention in Eqn.~(\ref{eq:attn}), which is parameter-free.
Thus, the DCAMA process is actually $n$-agnostic and can be applied for inference with arbitrary $n$.\footnote{Despite that, it is still computationally prohibitive to \textit{train} $n$-specific models for $n>1$, given both time and GPU memory considerations.}
Intuitively, the label of a query pixel is jointly determined by all available support pixels at once, irrespective of the exact number of support images.
This one-pass inference is distinct from the ensemble of individual inferences, where image-level predictions are first obtained with each support image independently and then merged to produce a set-level final prediction.
It is also different from some prototype-based $n$-shot approaches~\cite{dong2018few,yang2020prototype}, where features of all support images were processed simultaneously but compressed into one or few prototypes, losing pixel-level granularity.
Lastly, a minor adaption is to max-pool the skip connected support features across the support images, such that the entire DCAMA framework shown in Fig. \ref{fig:overview} becomes applicable to generic $n$-shot settings, where $n\geq1$.

\section{Experiments and Results}
\label{sec:exper}

\textbf{Datasets and evaluation.}
We evaluate the proposed method on three standard FSS benchmarks: PASCAL-5$^i$~\cite{shaban2017one}, COCO-20$^i$~\cite{nguyen2019feature} and FSS-1000~\cite{li2020fss}.
PASCAL-5$^i$ is created from PASCAL VOC 2012~\cite{everingham2010pascal}
and SDS~\cite{hariharan2014simultaneous} datasets.
It contains 20 classes, which are evenly split into four folds, i.e., five classes per fold.
COCO-20$^i$ is a larger and more challenging benchmark created from the COCO~\cite{lin2014microsoft} dataset.
It includes 80 classes, again evenly divided into four folds.
For both PASCAL-5$^i$ and COCO-20$^i$, the evaluation is done by cross-validation, where each fold is selected in turn as $\mathcal{D}_\mathrm{test}$, with the other three folds as $\mathcal{D}_\mathrm{train}$;
1,000 testing episodes are randomly sampled from $\mathcal{D}_\mathrm{test}$ for evaluation~\cite{min2021hypercorrelation}.
FSS-1000 \cite{li2020fss} comprises 1,000 classes that are divided into training, validation, and testing splits of 520, 240, and 240 classes, respectively, with 2,400 testing episodes sampled from the testing split for evaluation~\cite{min2021hypercorrelation}.
For the metrics, we adopt mean intersection over union (mIoU) and foreground-background IoU (FB-IoU)~\cite{min2021hypercorrelation}.
For PASCAL-5$^i$ and COCO-20$^i$, the mIoUs on individual folds, and the averaged mIoU and FB-IoU across the folds are reported;
for FSS-1000, the mIoU and FB-IoU on the testing split are reported.
Note that we attempt 
to follow the common practices adopted by previous works~\cite{min2021hypercorrelation,nguyen2019feature,tian2020prior,zhang2021few}
for fair comparisons.

\textbf{Implementation details.}
All experiments are conducted with the PyTorch \cite{paszke2019pytorch} framework (1.5.0).
{\color{blue}For the backbone feature extractor, we employ ResNet-50 and ResNet-101 \cite{he2016deep} pretrained on ImageNet \cite{deng2009imagenet} for their prevalent adoption in previous works.\footnote{\color{red}While several works (e.g., \cite{tian2020prior}) reported superior performance of ResNet-50 and ResNet-101, VGG-16 \cite{simonyan2014very} mostly produced inferior performance to the ResNet family in previous works on FSS.
Therefore, we do not include VGG-16 in our experiment.}
In addition, we also experiment with the base Swin Transformer model (Swin-B) pretrained on ImageNet-1K~\cite{liu2021swin}, to evaluate the generalization of our method on the non-convolutional backbone.}
We use three multi-layer DCAMA blocks for $\frac{1}{8}$, $\frac{1}{16}$, and $\frac{1}{32}$ scales, respectively, resulting in three pyramidal levels of cross attention weighted mask aggregation.
Unless otherwise specified, the last-layer features of the $\frac{1}{4}$ and $\frac{1}{8}$ scales are skip connected.
The input size of both the support and query images is $384 \times 384$ pixels.
{\color{purple}The mean binary cross-entropy loss is employed: $\mathcal{L}_\mathrm{BCE}=-\frac{1}{N}\sum[y\log p+(1-y)\log(1-p)]$, where $N$ is the total number of pixels, $y\in\{0,1\}$ is the pixel label (0 for background and 1 for foreground), and $p$ is the predicted probability.
We impose $\mathcal{L}_\mathrm{BCE}$ only on the final output to train our model, with the backbone parameters frozen.}
The SGD optimizer is employed, with the learning rate, momentum, and weight decay set to $10^{-3}$, 0.9, and $10^{-4}$, respectively.
The batch size is set to 48, 48, and 40 for PASCAL-5$^i$, COCO-20$^i$, and FSS-1000, respectively.
We follow HSNet~\cite{min2021hypercorrelation} to train our models without data augmentation and until convergence, for a fair comparison with the previous best performing method.
The training is done on four NVIDIA Tesla V100 GPUs and the inference is on an NVIDIA Tesla T4 GPU.
Our code is available at \url{https://github.com/pawn-sxy/DCAMA.git}.

\begin{table*}[t]\color{blue}
\caption{Performance on PASCAL-5$^i$ (top) and COCO-20$^i$ (bottom).
  HSNet$^\dagger$: our reimplementation based on official codes;
  $^\ast$: methods implementing the dot-product attention~\cite{vaswani2017attention}. 
  Bold and underlined numbers highlight the best and second best performance (if necessary) for each backbone, respectively.}\label{tab:pascal_n_coco}
  \centering
  \resizebox{\textwidth}{!}{
\begin{tabular}{ccccccccccccccc}
\hline
\rowcolor[HTML]{EFEFEF}
\multicolumn{15}{c}{\textit{PASCAL-5}$^i$~\cite{shaban2017one}}                                                                                                                                                                                                                                                                                                                                                                                                                                \\ \hline
\multicolumn{1}{c|}{\multirow{2}{*}{Backbone}}   & \multicolumn{1}{c|}{\multirow{2}{*}{Methods}}                       & \multicolumn{1}{c|}{\multirow{2}{*}{Type}}       & \multicolumn{6}{c|}{1-shot}                                                                                                                                        & \multicolumn{6}{c}{5-shot}                                                                                                                    \\ \cline{4-15}
\multicolumn{1}{c|}{}                            & \multicolumn{1}{c|}{}                                               & \multicolumn{1}{c|}{}                            & Fold-0        & Fold-1        & Fold-2        & \multicolumn{1}{c|}{Fold-3}        & \multicolumn{1}{c|}{mIoU}                & \multicolumn{1}{c|}{FB-IoU}        & Fold-0        & Fold-1        & Fold-2        & \multicolumn{1}{c|}{Fold-3}        & \multicolumn{1}{c|}{mIoU}                & FB-IoU        \\ \hline
\multicolumn{1}{c|}{\multirow{8}{*}{ResNet-50}}  & \multicolumn{1}{c|}{PPNet~\cite{liu2020part}}                       & \multicolumn{1}{c|}{\multirow{7}{*}{Prototype}}  & 52.7          & 62.8          & 57.4          & \multicolumn{1}{c|}{47.7}          & \multicolumn{1}{c|}{55.2 (5.6)}          & \multicolumn{1}{c|}{-}             & 60.3          & 70.0          & \uline{69.4}  & \multicolumn{1}{c|}{60.7}          & \multicolumn{1}{c|}{65.1 (4.6)}          & -             \\
\multicolumn{1}{c|}{}                            & \multicolumn{1}{c|}{PMM~\cite{yang2020prototype}}                   & \multicolumn{1}{c|}{}                            & 52.0          & 67.5          & 51.5          & \multicolumn{1}{c|}{49.8}          & \multicolumn{1}{c|}{55.2 (7.2)}          & \multicolumn{1}{c|}{-}             & 55.0          & 68.2          & 52.9          & \multicolumn{1}{c|}{51.1}          & \multicolumn{1}{c|}{56.8 (6.7)}          & -             \\
\multicolumn{1}{c|}{}                            & \multicolumn{1}{c|}{RPMM~\cite{yang2020prototype}}                  & \multicolumn{1}{c|}{}                            & 55.2          & 66.9          & 52.6          & \multicolumn{1}{c|}{50.7}          & \multicolumn{1}{c|}{56.3 (6.3)}          & \multicolumn{1}{c|}{-}             & 56.3          & 67.3          & 54.5          & \multicolumn{1}{c|}{51.0}          & \multicolumn{1}{c|}{57.3 (6.1)}          & -             \\
\multicolumn{1}{c|}{}                            & \multicolumn{1}{c|}{RePRI~\cite{boudiaf2021few}}                    & \multicolumn{1}{c|}{}                            & 59.8          & 68.3          & \textbf{62.1} & \multicolumn{1}{c|}{48.5}          & \multicolumn{1}{c|}{59.7 (7.2)}          & \multicolumn{1}{c|}{-}             & 64.6          & \uline{71.4}  & \textbf{71.1} & \multicolumn{1}{c|}{59.3}          & \multicolumn{1}{c|}{\uline{66.6} (5.0)}  & -             \\
\multicolumn{1}{c|}{}                            & \multicolumn{1}{c|}{PEFNet~\cite{tian2020prior}}                    & \multicolumn{1}{c|}{}                            & 61.7          & 69.5          & 55.4          & \multicolumn{1}{c|}{56.3}          & \multicolumn{1}{c|}{60.8 (5.6)}          & \multicolumn{1}{c|}{\uline{73.3}}  & 63.1          & 70.7          & 55.8          & \multicolumn{1}{c|}{57.9}          & \multicolumn{1}{c|}{61.9 (5.7)}          & \uline{73.9}  \\
\multicolumn{1}{c|}{}                            & \multicolumn{1}{c|}{SCL~\cite{zhang2021self}}                       & \multicolumn{1}{c|}{}                            & \uline{63.0}  & 70.0          & 56.5          & \multicolumn{1}{c|}{\uline{57.7}}  & \multicolumn{1}{c|}{61.8 (5.3)}          & \multicolumn{1}{c|}{71.9}          & 64.5          & 70.9          & 57.3          & \multicolumn{1}{c|}{58.7}          & \multicolumn{1}{c|}{62.9 (5.4)}          & 72.8          \\
\multicolumn{1}{c|}{}                            & \multicolumn{1}{c|}{TRFS$^\ast$~\cite{sun2021boosting}}             & \multicolumn{1}{c|}{}                            & 62.9          & \uline{70.7}  & 56.5          & \multicolumn{1}{c|}{57.5}          & \multicolumn{1}{c|}{\uline{61.9} (5.6)}  & \multicolumn{1}{c|}{-}             & \uline{65.0}  & 71.2          & 55.5          & \multicolumn{1}{c|}{\uline{60.9}}  & \multicolumn{1}{c|}{63.2 (5.7)}          & -             \\ \cline{2-15}
\multicolumn{1}{c|}{}                            & \multicolumn{1}{c|}{DCAMA (Ours)$^\ast$}                            & \multicolumn{1}{c|}{Pixel-wise}                  & \textbf{67.5} & \textbf{72.3} & \uline{59.6}  & \multicolumn{1}{c|}{\textbf{59.0}} & \multicolumn{1}{c|}{\textbf{64.6} (5.6)} & \multicolumn{1}{c|}{\textbf{75.7}} & \textbf{70.5} & \textbf{73.9} & 63.7          & \multicolumn{1}{c|}{\textbf{65.8}} & \multicolumn{1}{c|}{\textbf{68.5} (4.0)} & \textbf{79.5} \\ \hline
\multicolumn{1}{c|}{\multirow{6}{*}{ResNet-101}} & \multicolumn{1}{c|}{CWT$^\ast$~\cite{lu2021simpler}}                & \multicolumn{1}{c|}{\multirow{2}{*}{Prototype}}  & 56.9          & 65.2          & 61.2          & \multicolumn{1}{c|}{48.8}          & \multicolumn{1}{c|}{58.0 (6.1)}          & \multicolumn{1}{c|}{-}             & 62.6          & 70.2          & \textbf{68.8}  & \multicolumn{1}{c|}{57.2}          & \multicolumn{1}{c|}{64.7 (5.2)}          & -             \\
\multicolumn{1}{c|}{}                            & \multicolumn{1}{c|}{DoG-LSTM~\cite{azad2021texture}}                & \multicolumn{1}{c|}{}                            & 57.0          & 67.2          & 56.1          & \multicolumn{1}{c|}{54.3}          & \multicolumn{1}{c|}{58.7 (5.0)}          & \multicolumn{1}{c|}{-}             & 57.3          & 68.5          & 61.5          & \multicolumn{1}{c|}{56.3}          & \multicolumn{1}{c|}{60.9 (4.8)}          & -             \\ \cline{2-15}
\multicolumn{1}{c|}{}                            & \multicolumn{1}{c|}{DAN~\cite{wang2020few}}                         & \multicolumn{1}{c|}{\multirow{4}{*}{Pixel-wise}} & 54.7          & 68.6          & 57.8          & \multicolumn{1}{c|}{51.6}          & \multicolumn{1}{c|}{58.2 (6.4)}          & \multicolumn{1}{c|}{\uline{71.9}}          & 57.9          & 69.0          & 60.1          & \multicolumn{1}{c|}{54.9}          & \multicolumn{1}{c|}{60.5 (5.3)}          & 72.3          \\
\multicolumn{1}{c|}{}                            & \multicolumn{1}{c|}{CyCTR$^\ast$~\cite{zhang2021few}}               & \multicolumn{1}{c|}{}                            & \textbf{69.3} & \textbf{72.7} & 56.5          & \multicolumn{1}{c|}{\uline{58.6}}  & \multicolumn{1}{c|}{64.3 (6.9)}          & \multicolumn{1}{c|}{-}             & \textbf{73.5} & \uline{74.0}  & 58.6          & \multicolumn{1}{c|}{60.2}          & \multicolumn{1}{c|}{66.6 (7.2)}          & -             \\
\multicolumn{1}{c|}{}                            & \multicolumn{1}{c|}{HSNet~\cite{min2021hypercorrelation}}           & \multicolumn{1}{c|}{}                            & \uline{67.3}  & \uline{72.3}  & \uline{62.0}  & \multicolumn{1}{c|}{\textbf{63.1}} & \multicolumn{1}{c|}{\textbf{66.2} (4.1)} & \multicolumn{1}{c|}{\textbf{77.6}} & \uline{71.8}  & \textbf{74.4} & \uline{67.0}          & \multicolumn{1}{c|}{\textbf{68.3}} & \multicolumn{1}{c|}{\textbf{70.4} (2.9)} & \uline{80.6}  \\
\multicolumn{1}{c|}{}                            & \multicolumn{1}{c|}{DCAMA (Ours)$^\ast$}                            & \multicolumn{1}{c|}{}                            & 65.4          & 71.4          & \textbf{63.2} & \multicolumn{1}{c|}{58.3}          & \multicolumn{1}{c|}{\uline{64.6} (4.7)}  & \multicolumn{1}{c|}{\textbf{77.6}}  & 70.7          & 73.7          & 66.8 & \multicolumn{1}{c|}{\uline{61.9}}  & \multicolumn{1}{c|}{\uline{68.3} (4.4)}  & \textbf{80.8} \\ \hline
\multicolumn{1}{c|}{\multirow{2}{*}{Swin-B}}     & \multicolumn{1}{c|}{HSNet$^\dagger$~\cite{min2021hypercorrelation}} & \multicolumn{1}{c|}{\multirow{2}{*}{Pixel-wise}} & 67.9          & \textbf{74.0} & 60.3          & \multicolumn{1}{c|}{67.0}          & \multicolumn{1}{c|}{67.3 (4.9)}          & \multicolumn{1}{c|}{77.9}          & 72.2          & \textbf{77.5} & 64.0          & \multicolumn{1}{c|}{72.6}          & \multicolumn{1}{c|}{71.6 (4.8)}          & 81.2          \\
\multicolumn{1}{c|}{}                            & \multicolumn{1}{c|}{DCAMA (Ours)$^\ast$}                            & \multicolumn{1}{c|}{}                            & \textbf{72.2} & 73.8          & \textbf{64.3} & \multicolumn{1}{c|}{\textbf{67.1}} & \multicolumn{1}{c|}{\textbf{69.3} (3.8)} & \multicolumn{1}{c|}{\textbf{78.5}} & \textbf{75.7} & 77.1          & \textbf{72.0} & \multicolumn{1}{c|}{\textbf{74.8}} & \multicolumn{1}{c|}{\textbf{74.9} (1.8)} & \textbf{82.9} \\ \hline
\rowcolor[HTML]{EFEFEF}
\multicolumn{15}{c}{\textit{COCO-20}$^i$~\cite{nguyen2019feature}}                                                                                                                                                                                                                                                                                                                                                                                                                             \\ \hline
\multicolumn{1}{c|}{\multirow{7}{*}{ResNet-50}}  & \multicolumn{1}{c|}{PPNet~\cite{liu2020part}}                       & \multicolumn{1}{c|}{\multirow{5}{*}{Prototype}}  & 36.5          & 26.5          & 26.0          & \multicolumn{1}{c|}{19.7}          & \multicolumn{1}{c|}{27.2 (6.0)}          & \multicolumn{1}{c|}{-}             & \textbf{48.9} & 31.4          & 36.0          & \multicolumn{1}{c|}{30.6}          & \multicolumn{1}{c|}{36.7 (7.3)}          & -             \\
\multicolumn{1}{c|}{}                            & \multicolumn{1}{c|}{PMM~\cite{yang2020prototype}}                   & \multicolumn{1}{c|}{}                            & 29.3          & 34.8          & 27.1          & \multicolumn{1}{c|}{27.3}          & \multicolumn{1}{c|}{29.6 (3.1)}          & \multicolumn{1}{c|}{-}             & 33.0          & 40.6          & 30.3          & \multicolumn{1}{c|}{33.3}          & \multicolumn{1}{c|}{34.3 (3.8)}          & -             \\
\multicolumn{1}{c|}{}                            & \multicolumn{1}{c|}{RPMM~\cite{yang2020prototype}}                  & \multicolumn{1}{c|}{}                            & 29.5          & 36.8          & 28.9          & \multicolumn{1}{c|}{27.0}          & \multicolumn{1}{c|}{30.6 (3.7)}          & \multicolumn{1}{c|}{-}             & 33.8          & 42.0          & 33.0          & \multicolumn{1}{c|}{33.3}          & \multicolumn{1}{c|}{35.5 (3.7)}          & -             \\
\multicolumn{1}{c|}{}                            & \multicolumn{1}{c|}{TRFS$^\ast$~\cite{sun2021boosting}}             & \multicolumn{1}{c|}{}                            & 31.8          & 34.9          & 36.4          & \multicolumn{1}{c|}{31.4}          & \multicolumn{1}{c|}{33.6 (2.1)}          & \multicolumn{1}{c|}{-}             & 35.4          & 41.7          & 42.3          & \multicolumn{1}{c|}{36.1}          & \multicolumn{1}{c|}{38.9 (3.1)}          & -             \\
\multicolumn{1}{c|}{}                            & \multicolumn{1}{c|}{RePRI~\cite{boudiaf2021few}}                    & \multicolumn{1}{c|}{}                            & 31.2          & 38.1          & 33.3          & \multicolumn{1}{c|}{33.0}          & \multicolumn{1}{c|}{34.0 (2.6)}          & \multicolumn{1}{c|}{-}             & 38.5          & 46.2          & 40.0          & \multicolumn{1}{c|}{43.6}          & \multicolumn{1}{c|}{42.1 (3.0)}          & -             \\ \cline{2-15}
\multicolumn{1}{c|}{}                            & \multicolumn{1}{c|}{CyCTR$^\ast$~\cite{zhang2021few}}               & \multicolumn{1}{c|}{\multirow{2}{*}{Pixel-wise}} & \uline{38.9}  & \uline{43.0}  & \uline{39.6}  & \multicolumn{1}{c|}{\uline{39.8}}  & \multicolumn{1}{c|}{\uline{40.3} (1.6)}  & \multicolumn{1}{c|}{-}             & 41.1          & \uline{48.9}  & \uline{45.2}  & \multicolumn{1}{c|}{\textbf{47.0}} & \multicolumn{1}{c|}{\uline{45.6} (2.9)}  & -             \\
\multicolumn{1}{c|}{}                            & \multicolumn{1}{c|}{DCAMA (Ours)$^\ast$}                            & \multicolumn{1}{c|}{}                            & \textbf{41.9} & \textbf{45.1} & \textbf{44.4} & \multicolumn{1}{c|}{\textbf{41.7}} & \multicolumn{1}{c|}{\textbf{43.3} (1.5)} & \multicolumn{1}{c|}{69.5}          & \uline{45.9}  & \textbf{50.5} & \textbf{50.7} & \multicolumn{1}{c|}{\uline{46.0}}  & \multicolumn{1}{c|}{\textbf{48.3} (2.3)} & 71.7          \\ \hline
\multicolumn{1}{c|}{\multirow{6}{*}{ResNet-101}} & \multicolumn{1}{c|}{CWT$^\ast$~\cite{lu2021simpler}}                & \multicolumn{1}{c|}{\multirow{3}{*}{Prototype}}  & 30.3          & 36.6          & 30.5          & \multicolumn{1}{c|}{32.2}          & \multicolumn{1}{c|}{32.4 (2.5)}          & \multicolumn{1}{c|}{-}             & 38.5          & 46.7          & 39.4          & \multicolumn{1}{c|}{43.2}          & \multicolumn{1}{c|}{42.0 (3.3)}          & -             \\
\multicolumn{1}{c|}{}                            & \multicolumn{1}{c|}{SCL~\cite{zhang2021self}}                       & \multicolumn{1}{c|}{}                            & 36.4          & 38.6          & 37.5          & \multicolumn{1}{c|}{35.4}          & \multicolumn{1}{c|}{37.0 (1.2)}          & \multicolumn{1}{c|}{-}             & 38.9          & 40.5          & 41.5          & \multicolumn{1}{c|}{38.7}          & \multicolumn{1}{c|}{39.9 (1.2)}          & -             \\
\multicolumn{1}{c|}{}                            & \multicolumn{1}{c|}{PEFNet~\cite{tian2020prior}}                    & \multicolumn{1}{c|}{}                            & 36.8          & 41.8          & 38.7          & \multicolumn{1}{c|}{36.7}          & \multicolumn{1}{c|}{38.5 (2.1)}          & \multicolumn{1}{c|}{63.0}          & 40.4          & 46.8          & 43.2          & \multicolumn{1}{c|}{40.5}          & \multicolumn{1}{c|}{42.7 (2.6)}          & 65.8          \\ \cline{2-15}
\multicolumn{1}{c|}{}                            & \multicolumn{1}{c|}{DAN~\cite{wang2020few}}                         & \multicolumn{1}{c|}{\multirow{3}{*}{Pixel-wise}} & -             & -             & -             & \multicolumn{1}{c|}{-}             & \multicolumn{1}{c|}{24.4 (-)}            & \multicolumn{1}{c|}{62.3}          & -             & -             & -             & \multicolumn{1}{c|}{-}             & \multicolumn{1}{c|}{29.6 (-)}            & 63.9          \\
\multicolumn{1}{c|}{}                            & \multicolumn{1}{c|}{HSNet~\cite{min2021hypercorrelation}}           & \multicolumn{1}{c|}{}                            & \uline{37.2}  & \uline{44.1}  & \uline{42.4}  & \multicolumn{1}{c|}{\textbf{41.3}} & \multicolumn{1}{c|}{\uline{41.2} (2.5)}  & \multicolumn{1}{c|}{\uline{69.1}}  & \uline{45.9}  & \uline{53.0}  & \uline{51.8}  & \multicolumn{1}{c|}{\textbf{47.1}} & \multicolumn{1}{c|}{\uline{49.5} (3.0)}  & \uline{72.4}  \\
\multicolumn{1}{c|}{}                            & \multicolumn{1}{c|}{DCAMA (Ours)$^\ast$}                            & \multicolumn{1}{c|}{}                            & \textbf{41.5} & \textbf{46.2} & \textbf{45.2} & \multicolumn{1}{c|}{\textbf{41.3}} & \multicolumn{1}{c|}{\textbf{43.5} (2.2)} & \multicolumn{1}{c|}{\textbf{69.9}} & \textbf{48.0} & \textbf{58.0} & \textbf{54.3} & \multicolumn{1}{c|}{\textbf{47.1}} & \multicolumn{1}{c|}{\textbf{51.9} (4.5)} & \textbf{73.3} \\ \hline
\multicolumn{1}{c|}{\multirow{2}{*}{Swin-B}}     & \multicolumn{1}{c|}{HSNet$^\dagger$~\cite{min2021hypercorrelation}} & \multicolumn{1}{c|}{\multirow{2}{*}{Pixel-wise}} & 43.6          & 49.9          & 49.4          & \multicolumn{1}{c|}{46.4}          & \multicolumn{1}{c|}{47.3 (2.5)}          & \multicolumn{1}{c|}{72.5}          & 50.1          & 58.6          & 56.7          & \multicolumn{1}{c|}{55.1}          & \multicolumn{1}{c|}{55.1 (3.2)}          & 76.1          \\
\multicolumn{1}{c|}{}                            & \multicolumn{1}{c|}{DCAMA (Ours)$^\ast$}                            & \multicolumn{1}{c|}{}                            & \textbf{49.5} & \textbf{52.7} & \textbf{52.8} & \multicolumn{1}{c|}{\textbf{48.7}} & \multicolumn{1}{c|}{\textbf{50.9} (1.8)} & \multicolumn{1}{c|}{\textbf{73.2}} & \textbf{55.4} & \textbf{60.3} & \textbf{59.9} & \multicolumn{1}{c|}{\textbf{57.5}} & \multicolumn{1}{c|}{\textbf{58.3} (2.0)} & \textbf{76.9} \\ \hline
\end{tabular}
  }\vspace{-3.6mm}
\end{table*}

\begin{table}[t]\color{blue}
  \caption{Performance on FSS-1000~\cite{li2020fss}.
  HSNet$^\dagger$: our reimplementation based on the official codes.
  Bold and underlined numbers highlight the best and second best performance (if necessary) for each backbone, respectively.}\label{tab:fss}
  \centering
  \resizebox{.6\linewidth}{!}{
\begin{tabular}{c|c|c|cc|cc}
\hline
\multirow{2}{*}{Backbone}   & \multirow{2}{*}{Methods}                       & \multirow{2}{*}{Type}       & \multicolumn{2}{c|}{1-shot}   & \multicolumn{2}{c}{5-shot}    \\ \cline{4-7}
                            &                                                &                             & mIoU          & FB-IoU        & mIoU          & FB-IoU        \\ \hline
ResNet-50                   & DCAMA (Ours)                                   & Pixel-wise                  & 88.2          & 92.5          & 88.8          & 92.9          \\ \hline
\multirow{4}{*}{ResNet-101} & DoG-LSTM~\cite{azad2021texture}                & Prototype                   & 80.8          & -             & 83.4          & -             \\ \cline{2-7}
                            & DAN~\cite{wang2020few}                         & \multirow{3}{*}{Pixel-wise} & 85.2          & -             & 88.1          & -             \\
                            & HSNet~\cite{min2021hypercorrelation}           &                             & \uline{86.5}  & -             & \uline{88.5}  & -             \\
                            & DCAMA (Ours)                                   &                             & \textbf{88.3} & 92.4          & \textbf{89.1} & 93.1          \\ \hline
\multirow{2}{*}{Swin-B}     & HSNet$^\dagger$~\cite{min2021hypercorrelation} & \multirow{2}{*}{Pixel-wise} & 86.7          & 91.8          & 88.9          & 93.2          \\
                            & DCAMA (Ours)                                   &                             & \textbf{90.1} & \textbf{93.8} & \textbf{90.4} & \textbf{94.1} \\ \hline
\end{tabular}
  }
\end{table}

\subsection{Comparison with State of the Art}
In Table~\ref{tab:pascal_n_coco} and Table~\ref{tab:fss}, we compare the performance of our proposed DCAMA framework with that of SOTA approaches to FSS published since 2020 on PASCAL-5$^i$, COCO-20$^i$, and FSS-1000, respectively.
Unless otherwise specified, reported numbers of other methods are from the original papers;
when results with different backbones are available, we report the higher ones only to save space.

\begin{figure*}[!t]
  \centering
  \includegraphics[trim=0 8 0 0, clip, width=\textwidth]{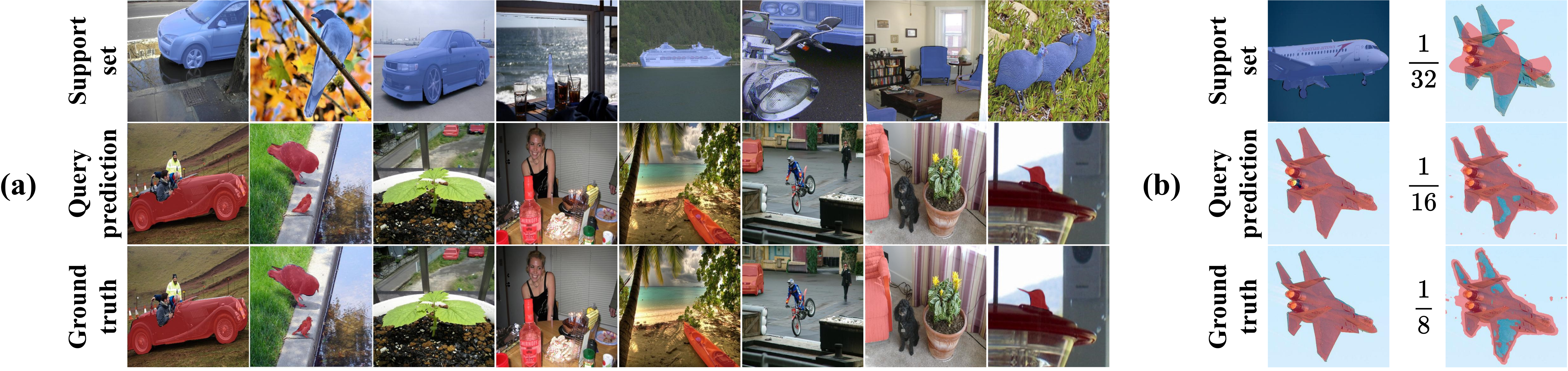}
  \caption{(a) Qualitative results on PASCAL-5$^i$ in 1-shot setting, in the presence of intra-class variations, size differences, complex background, and occlusions.
  (b) Multi-scale intermediate query masks aggregated by the multi-layer DCAMA blocks for a 1-shot task sampled from PASCAL-5$^i$.}
  \label{fig:qualitativeresults}
\end{figure*}

Above all, our method
{\color{blue} is very competitive: it achieves the best performance in terms of both mIoU and FB-IoU for almost all combinations of the backbone networks (ResNet-50, ResNet-101, and Swin-B) and few-shot settings (1- and 5-shot) on all the three benchmark datasets.
The only exception is on PASCAL-5\textsuperscript{i} with ResNet-101, where {\color{red}our DCAMA and HSNet \cite{min2021hypercorrelation} share the top two spots for mIoU and FB-IoU in 1- and 5-shot settings with comparable performance}.
When employing Swin-B as the backbone feature extractor, DCAMA significantly advances the system-level SOTA on all the three benchmarks upon the previous three-benchmark SOTA HSNet (ResNet-101), e.g., by 3.1\% (1-shot) and 4.5\% (5-shot) on PASCAL-5$^i$, 9.7\% and 8.8\% on COCO-20$^i$, and 3.6\% and 1.9\% on FSS-1000, respectively, in terms of mIoU.
Second,
although the performance of HSNet improves with the Swin-B backbone,
yet still suffers considerable disadvantages of 2--3.6\% (1-shot) and 1.5--3.3\% (5-shot) in mIoU from that of DCAMA.
In addition, our DCAMA is also more stable than HSNet across the folds, observing the lower standard deviations.
These results demonstrate DCAMA's applicability to both convolution- and attention-based backbones.}
Third, the performance of the methods based on pixel-wise correlations (except for DAN~\cite{wang2020few}) is generally better than that of those based on prototyping, confirming the intuition of making use of the fine-grained pixel-level information for the task of FSS.
{\color{blue}Last but not least, it is worth noting that our DCAMA demonstrates consistent advantages over the other three methods (CWT~\cite{lu2021simpler}, TRFS~\cite{sun2021boosting}, and CyCTR~\cite{zhang2021few}) that also implemented the dot-product attention of the Transformer~\cite{vaswani2017attention}.}
Fig. \ref{fig:qualitativeresults}(a) visualizes some segmentation results by DCAMA in challenging cases.
More results and visualizations, including region-wise over- and under-segmentation measures \cite{zhang2021rethinking}, are given in the supplementary material.

\textit{Remarks.}
Although the three best performing methods in Table~\ref{tab:pascal_n_coco} (HSNet~\cite{min2021hypercorrelation}, CyCTR~\cite{zhang2021few}, and ours) all rely on pixel-level cross-query-and-support similarities, their underlying concepts of query label inference are quite different and worth clarification.
HSNet predicted the query label based on the similarities of a query pixel to all the foreground support pixels (while ignoring the background);
intuitively, the more similar a query pixel is to the foreground support pixels, the more likely it is foreground.
CyCTR first reconstructed the query features from the support features based on the similarities to subsets of both foreground and background support pixels, then trained a classifier on the reconstructed query features.
Our DCAMA directly aggregates the query label from the support mask weighted by the query pixel's similarities to all support pixels, representing a totally different concept.

\begin{figure}[!t]
  \centering
   \includegraphics[trim=0 0 0 0, clip, width=.249\linewidth]{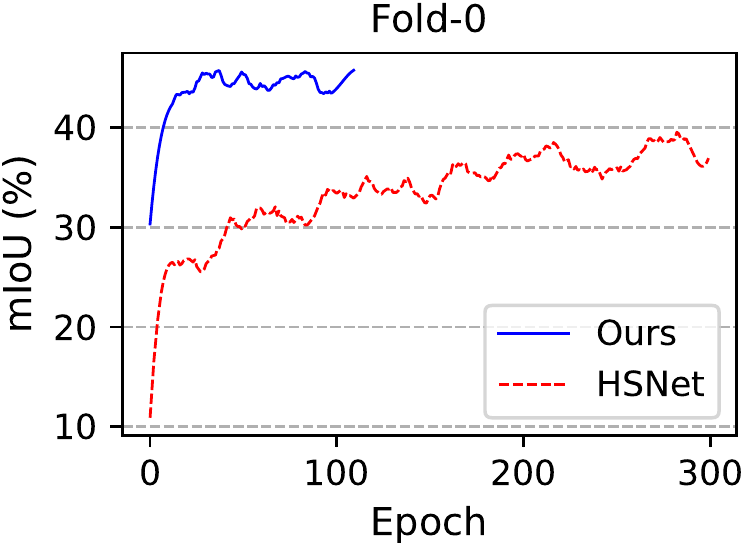}\hfill%
   \includegraphics[trim=15 0 0 0, clip, width=.23\linewidth]{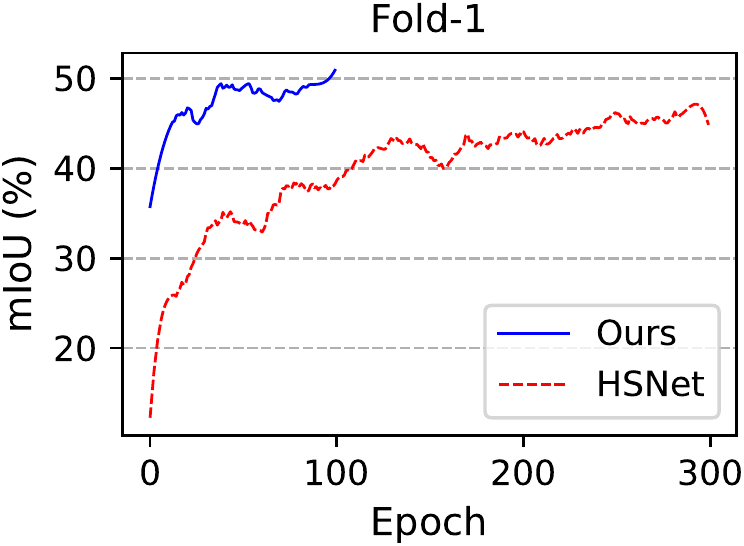}\hfill%
   \includegraphics[trim=15 0 0 0, clip, width=.23\linewidth]{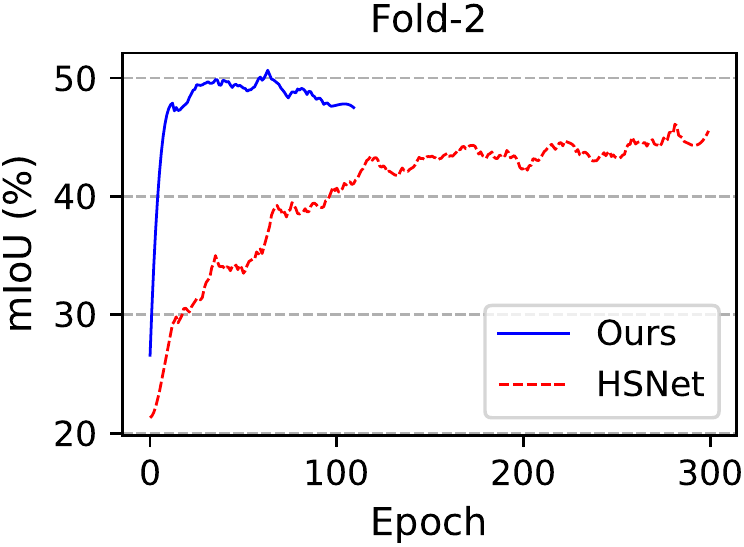}\hfill%
   \includegraphics[trim=15 0 0 0, clip, width=.23\linewidth]{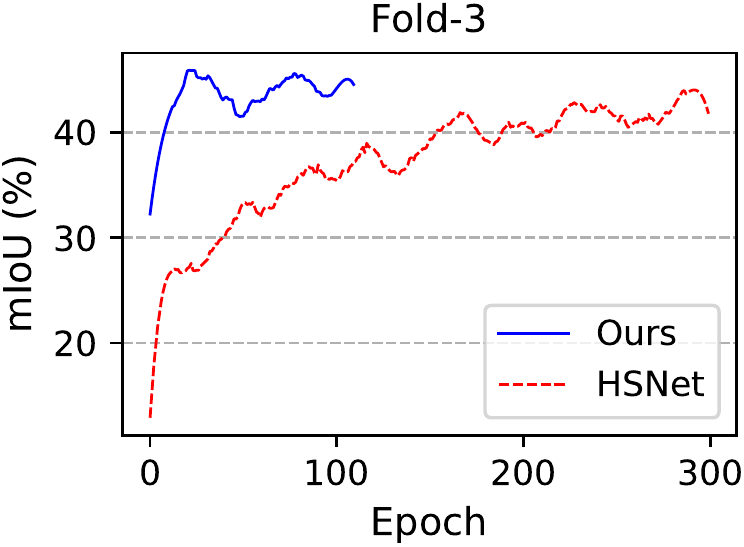}
   \caption{The mIoU curves on the validation set during training on COCO-20$^i$.
   The curves of HSNet~\cite{min2021hypercorrelation} are produced with the official codes released by the authors.}
   \label{fig:traincurve}
\end{figure}

\textbf{Training {\color{blue}and inference} efficiency.}
We compare the training efficiency of our method to that of HSNet~\cite{min2021hypercorrelation}.
As both methods incur the computational complexity of $O(N^2)$ for pixel-wise correlation,
they also take comparable amounts of time for training per epoch (e.g., around four minutes
with our hardware and training setting on COCO-20$^i$).
However, as shown in Fig. \ref{fig:traincurve}, our method takes much fewer training epochs to converge.
Therefore, our DCAMA is also more efficient in terms of training time, in addition to achieving higher performance than the previous SOTA method HSNet.
{\color{blue}As to inference, DCAMA runs at comparable speed with HSNet using the same backbone, e.g., about 8 and 20 frames per second (FPS) with Swin-B and ResNet-101, respectively, for 1-shot segmentation on an entry-level NVIDIA Tesla T4 GPU.
In contrast, CyCTR \cite{zhang2021few} runs at about three FPS with the ResNet-50 backbone.}


\subsection{Ablation Study}
We conduct thorough ablation studies on the PASCAL-5$^i$~\cite{shaban2017one} dataset, to gain deeper understanding of our proposed DCAMA framework and verify its design. {\color{blue}Swin-B~\cite{liu2021swin} is used as the backbone feature extractor for the ablation studies.}

\begin{figure}[t]
  \centering
   \includegraphics[trim=0 0 0 0, clip, width=.85\linewidth]{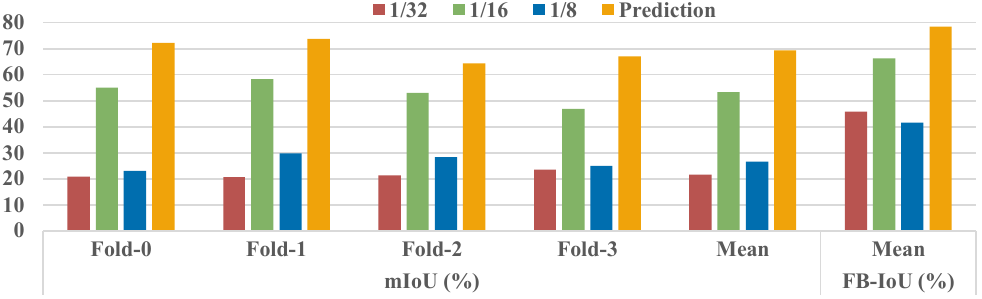}
   \caption{Performance of the multi-scale intermediate query masks aggregated by the multi-layer DCAMA blocks and the final prediction (1-shot on PASCAL-5$^i$).}
   \label{fig:middle}
\end{figure}

\textbf{Intermediate query masks aggregated by the multi-layer DCAMA blocks.}
We first validate the physical meaning of the proposed mask aggregation paradigm, by verifying that the outputs of the multi-layer DCAMA blocks (Fig.~\ref{fig:nshot}) are indeed meaningful segmentations.
For this purpose, we sum $\hat{M}^q_i$ along the layer dimension for scales $i\in\{\frac{1}{8},\frac{1}{16},\frac{1}{32}\}$, binarize the sum with the Otsu's method~\cite{otsu1979threshold}, and resize the resulting masks to evaluate against the ground truth for 1-shot segmentation.
As shown in Fig. \ref{fig:middle}, the mIoU and FB-IoU of the $\frac{1}{16}$ scale masks are fairly high, close to those of some comparison methods in Table~\ref{tab:pascal_n_coco}.
Meanwhile, those of the $\frac{1}{8}$ and $\frac{1}{32}$ scale masks are much lower, which may be because: the $\frac{1}{8}$ scale features have not learned enough high-level semantics, and the $\frac{1}{32}$ scale features are too abstract/class-specific and coarse.
The final prediction effectively integrates the multi-scale intermediate masks and achieves optimal performance.
For intuitive perception, we also visualize the intermediate masks for a specific 1-shot task in Fig. \ref{fig:qualitativeresults}(b), where the overlays clearly demonstrate that they are valid segmentations, albeit less accurate than the final prediction.
These results verify that the proposed DCAMA indeed functions as designed, and that the multi-scale strategy is effective.

\textbf{Effect of background support information.}
A notable difference between our DCAMA and the previous SOTA HSNet~\cite{min2021hypercorrelation} is that HSNet ignored the background support features whereas DCAMA uses them for full utilization of the available information.
To evaluate the actual effect of the difference, we conduct an ablative experiment where the background pixels are zeroed out from the support feature maps before fed to the multi-layer DCAMA blocks {\color{blue}for training and inference}---similar to HSNet, for 1-shot segmentation.
The results in Table~\ref{tab:maskground} show that ignoring background support information leads to decreases of 2.0\% and 2.3\% in mIoU and FB-IoU, respectively, suggesting the value of fully utilizing all the information available in the support set.

\begin{table}[t]\vspace{-7.2mm}
\centering
\begin{minipage}[t]{0.455\textwidth}
    \begin{table}[H]
        \setlength{\belowcaptionskip}{2mm}%
        \caption{Ablation study on the effect of the background support information (1-shot on PASCAL-5$^i$).}\label{tab:maskground}
        \centering
        \begin{adjustbox}{width=\linewidth}
            \begin{tabular}{c|cccc|c|c}
            \hline
            Background & Fold-0 & Fold-1 & Fold-2 & Fold-3 & mIoU  & FB-IoU \\
            \hline
            \checkmark & 72.2  & \textbf{73.8} & \textbf{64.3} & 67.1  & \textbf{69.3} & \textbf{78.5 } \\
            \ding{55} & \textbf{73.3} & 72.7  & 53.6  & \textbf{69.8} & 67.3  & 76.2  \\
            \hline
            \end{tabular}%
        \end{adjustbox}
    \end{table}
\end{minipage}\hfill%
\begin{minipage}[t]{.5\textwidth}
  \begin{table}[H]
      \setlength{\belowcaptionskip}{2mm}%
      \caption{Ablation study on strategies for $n$-shot segmentation (5-shot on PASCAL-5$^i$).}\label{tab:kshotresults}
      \centering
      \setlength{\tabcolsep}{1.3mm}
      \begin{adjustbox}{width=\linewidth}
        \begin{tabular}{c|cccc|c}
        \hline
        Strategy & Voting & Averaging & HSNet~\cite{min2021hypercorrelation} & SCL~\cite{zhang2021self} & Ours \\
        \hline
        mIoU  & 74.0  & 74.0  & 73.9  & 74.1  & \textbf{74.9} \\
        FB-IoU & 82.0  & 82.0  & 81.8  & 82.0  & \textbf{82.9} \\
        \hline
        \end{tabular}%
      \end{adjustbox}
  \end{table}
\end{minipage}\vspace{-2mm}%
\end{table}





\textbf{Strategy for \textit{n}-shot inference.}
To verify the effectiveness of our proposed one-pass $n$-shot inference, we compare its 5-shot performance to the following ensembles of five 1-shot predictions:
naive voting and averaging, the normalized voting in HSNet~\cite{min2021hypercorrelation}, and cross-guided averaging in SCL~\cite{zhang2021self}.
As shown in Table~\ref{tab:kshotresults}, our strategy outperforms all the
ensembles, suggesting the advantage in collectively utilizing all available support features together
for FSS.

\section{Conclusion}
\label{sec:con}
In this work, we proposed a new paradigm based on metric learning for Few-shot Semantic Segmentation (FSS): Dense Cross-query-and-support Attention weighted Mask Aggregation (DCAMA).
In addition, we implemented the DCAMA framework with the scaled dot-production attention in the Transformer structure, for simplicity and efficiency.
The DCAMA framework was distinct from previous works in three aspects: (i) it directly predicted the mask value of a query pixel as an additive aggregation of the mask values of all support pixels, weighted by pixel-wise similarities between the query and support features;
(ii) it fully utilized all support pixels, including both foreground and background; and
(iii) it proposed efficient and effective one-pass $n$-shot inference which considered pixels from all support images simultaneously.
Experiments showed that our DCAMA framework set the new state of the art on all three commonly used FSS benchmarks.
For future research, it is tempting to adapt the paradigm for other few-shot learning tasks that involve dense predictions such as detection.

\subsubsection{Acknowledgement.} This work was supported by the National Key
R\&D Program of China (2018AAA0100104, 2018AAA0100100), Natural Science Foundation of Jiangsu Province (BK20211164).

%
%
\bibliographystyle{splncs04}
\bibliography{egbib}

\begin{thebibliography}{10}
\providecommand{\url}[1]{\texttt{#1}}
\providecommand{\urlprefix}{URL }
\providecommand{\doi}[1]{https://doi.org/#1}

\bibitem{azad2021texture}
Azad, R., Fayjie, A.R., Kauffmann, C., Ben~Ayed, I., Pedersoli, M., Dolz, J.:
  On the texture bias for few-shot {CNN} segmentation. In: Proceedings of the
  IEEE/CVF Winter Conference on Applications of Computer Vision. pp. 2674--2683
  (2021)

\bibitem{boudiaf2021few}
Boudiaf, M., Kervadec, H., Masud, Z.I., Piantanida, P., Ben~Ayed, I., Dolz, J.:
  Few-shot segmentation without meta-learning: A good transductive inference is
  all you need? In: Proceedings of the IEEE/CVF Conference on Computer Vision
  and Pattern Recognition. pp. 13979--13988 (2021)

\bibitem{chen2017deeplab}
Chen, L.C., Papandreou, G., Kokkinos, I., Murphy, K., Yuille, A.L.: Deep{L}ab:
  Semantic image segmentation with deep convolutional nets, atrous convolution,
  and fully connected {CRF}s. IEEE Transactions on Pattern Analysis and Machine
  Intelligence  \textbf{40}(4),  834--848 (2017)

\bibitem{chen2018encoder}
Chen, L.C., Zhu, Y., Papandreou, G., Schroff, F., Adam, H.: Encoder-decoder
  with atrous separable convolution for semantic image segmentation. In:
  Proceedings of the European Conference on Computer Vision. pp. 801--818
  (2018)

\bibitem{cui2021unified}
Cui, H., Wei, D., Ma, K., Gu, S., Zheng, Y.: A unified framework for
  generalized low-shot medical image segmentation with scarce data. IEEE
  Transactions on Medical Imaging  \textbf{40}(10),  2656--2671 (2021)

\bibitem{deng2009imagenet}
Deng, J., Dong, W., Socher, R., Li, L.J., Li, K., Fei-Fei, L.: {ImageNet: A
  large-scale hierarchical image database}. In: 2009 IEEE Conference on
  Computer Vision and Pattern Recognition. pp. 248--255. Ieee (2009)

\bibitem{dong2018few}
Dong, N., Xing, E.P.: Few-shot semantic segmentation with prototype learning.
  In: British Machine Vision Conference. vol.~3 (2018)

\bibitem{dong2018fast}
Dong, X., Zhu, L., Zhang, D., Yang, Y., Wu, F.: Fast parameter adaptation for
  few-shot image captioning and visual question answering. In: Proceedings of
  the ACM International Conference on Multimedia. pp. 54--62 (2018)

\bibitem{dosovitskiy2020image}
Dosovitskiy, A., Beyer, L., Kolesnikov, A., Weissenborn, D., Zhai, X.,
  Unterthiner, T., Dehghani, M., Minderer, M., Heigold, G., Gelly, S.,
  Uszkoreit, J., Houlsby, N.: An image is worth 16x16 words: {T}ransformers for
  image recognition at scale. International Conference on Learning
  Representations  (2021)

\bibitem{everingham2010pascal}
Everingham, M., Van~Gool, L., Williams, C.K., Winn, J., Zisserman, A.: The
  {PASCAL} visual object classes ({VOC}) challenge. International Journal of
  Computer Vision  \textbf{88}(2),  303--338 (2010)

\bibitem{fei2006one}
Fei-Fei, L., Fergus, R., Perona, P.: One-shot learning of object categories.
  IEEE Transactions on Pattern Analysis and Machine Intelligence
  \textbf{28}(4),  594--611 (2006)

\bibitem{fink2005object}
Fink, M.: Object classification from a single example utilizing class relevance
  metrics. Advances in Neural Information Processing Systems  \textbf{17},
  449--456 (2005)

\bibitem{hariharan2014simultaneous}
Hariharan, B., Arbel{\'a}ez, P., Girshick, R., Malik, J.: Simultaneous
  detection and segmentation. In: Proceedings of the European Conference on
  Computer Vision. pp. 297--312. Springer (2014)

\bibitem{he2016deep}
He, K., Zhang, X., Ren, S., Sun, J.: Deep residual learning for image
  recognition. In: Proceedings of the IEEE Conference on Computer Vision and
  Pattern Recognition. pp. 770--778 (2016)

\bibitem{kulis2013metric}
Kulis, B., et~al.: Metric learning: A survey. Foundations and
  Trends{\textregistered} in Machine Learning  \textbf{5}(4),  287--364 (2013)

\bibitem{lake2011Omniglot}
Lake, B., Salakhutdinov, R., Gross, J., Tenenbaum, J.: One shot learning of
  simple visual concepts. In: Proceedings of the Annual Meeting of the
  Cognitive Science Society. vol.~33, pp. 2568--2573 (2011)

\bibitem{li2020fss}
Li, X., Wei, T., Chen, Y.P., Tai, Y.W., Tang, C.K.: {FSS}-1000: A 1000-class
  dataset for few-shot segmentation. In: Proceedings of the IEEE/CVF Conference
  on Computer Vision and Pattern Recognition. pp. 2869--2878 (2020)

\bibitem{lin2017feature}
Lin, T.Y., Doll{\'a}r, P., Girshick, R., He, K., Hariharan, B., Belongie, S.:
  Feature pyramid networks for object detection. In: Proceedings of the
  IEEE/CVF Conference on Computer Vision and Pattern Recognition. pp.
  2117--2125 (2017)

\bibitem{lin2014microsoft}
Lin, T.Y., Maire, M., Belongie, S., Hays, J., Perona, P., Ramanan, D.,
  Doll{\'a}r, P., Zitnick, C.L.: Microsoft {COCO}: Common objects in context.
  In: Proceedings of the European Conference on Computer Vision. pp. 740--755.
  Springer (2014)

\bibitem{liu2020part}
Liu, Y., Zhang, X., Zhang, S., He, X.: Part-aware prototype network for
  few-shot semantic segmentation. In: Proceedings of the European Conference on
  Computer Vision. pp. 142--158. Springer (2020)

\bibitem{liu2021swin}
Liu, Z., Lin, Y., Cao, Y., Hu, H., Wei, Y., Zhang, Z., Lin, S., Guo, B.: Swin
  {T}ransformer: Hierarchical vision {T}ransformer using shifted windows. In:
  Proceedings of the IEEE/CVF International Conference on Computer Vision. pp.
  10012--10022 (2021)

\bibitem{long2015fully}
Long, J., Shelhamer, E., Darrell, T.: Fully convolutional networks for semantic
  segmentation. In: Proceedings of the IEEE/CVF Conference on Computer Vision
  and Pattern Recognition. pp. 3431--3440 (2015)

\bibitem{lu2021simpler}
Lu, Z., He, S., Zhu, X., Zhang, L., Song, Y.Z., Xiang, T.: Simpler is better:
  Few-shot semantic segmentation with classifier weight {T}ransformer. In:
  Proceedings of the IEEE/CVF International Conference on Computer Vision. pp.
  8741--8750 (2021)

\bibitem{min2021hypercorrelation}
Min, J., Kang, D., Cho, M.: Hypercorrelation squeeze for few-shot segmentation.
  In: Proceedings of the IEEE/CVF International Conference on Computer Vision
  (2021)

\bibitem{minaee2021image}
Minaee, S., Boykov, Y.Y., Porikli, F., Plaza, A.J., Kehtarnavaz, N.,
  Terzopoulos, D.: Image segmentation using deep learning: A survey. IEEE
  Transactions on Pattern Analysis and Machine Intelligence pp.~1--1 (2021).
  \doi{10.1109/TPAMI.2021.3059968}

\bibitem{nguyen2019feature}
Nguyen, K., Todorovic, S.: Feature weighting and boosting for few-shot
  segmentation. In: Proceedings of the IEEE/CVF International Conference on
  Computer Vision. pp. 622--631 (2019)

\bibitem{otsu1979threshold}
Otsu, N.: A threshold selection method from gray-level histograms. IEEE
  Transactions on Systems, Man, and Cybernetics  \textbf{9}(1),  62--66 (1979)

\bibitem{paszke2019pytorch}
Paszke, A., Gross, S., Massa, F., Lerer, A., Bradbury, J., Chanan, G., Killeen,
  T., Lin, Z., Gimelshein, N., Antiga, L., et~al.: Py{T}orch: An imperative
  style, high-performance deep learning library. Advances in Neural Information
  Processing Systems  \textbf{32},  8026--8037 (2019)

\bibitem{ravi2016optimization}
Ravi, S., Larochelle, H.: Optimization as a model for few-shot learning. In:
  International Conference on Learning Representations (2017)

\bibitem{ronneberger2015u}
Ronneberger, O., Fischer, P., Brox, T.: U-{N}et: Convolutional networks for
  biomedical image segmentation. In: International Conference on Medical Image
  Computing and Computer-Assisted Intervention. pp. 234--241. Springer (2015)

\bibitem{shaban2017one}
Shaban, A., Bansal, S., Liu, Z., Essa, I., Boots, B.: One-shot learning for
  semantic segmentation. In: Proceedings of the British Machine Vision
  Conference. pp. 167.1--167.13 (2017)

\bibitem{simonyan2014very}
Simonyan, K., Zisserman, A.: Very deep convolutional networks for large-scale
  image recognition. arXiv preprint arXiv:1409.1556  (2014)

\bibitem{snell2017prototypical}
Snell, J., Swersky, K., Zemel, R.: Prototypical networks for few-shot learning.
  In: Advances in Neural Information Processing Systems. pp. 4080--4090 (2017)

\bibitem{strudel2021segmenter}
Strudel, R., Garcia, R., Laptev, I., Schmid, C.: Segmenter: {T}ransformer for
  semantic segmentation. In: Proceedings of the IEEE/CVF International
  Conference on Computer Vision. pp. 7262--7272 (2021)

\bibitem{sun2021boosting}
Sun, G., Liu, Y., Liang, J., Van~Gool, L.: Boosting few-shot semantic
  segmentation with {T}ransformers. arXiv preprint arXiv:2108.02266  (2021)

\bibitem{sung2018learning}
Sung, F., Yang, Y., Zhang, L., Xiang, T., Torr, P.H., Hospedales, T.M.:
  Learning to compare: Relation network for few-shot learning. In: Proceedings
  of the IEEE/CVF Conference on Computer Vision and Pattern Recognition. pp.
  1199--1208 (2018)

\bibitem{taghanaki2021deep}
Taghanaki, S.A., Abhishek, K., Cohen, J.P., Cohen-Adad, J., Hamarneh, G.: Deep
  semantic segmentation of natural and medical images: A review. Artificial
  Intelligence Review  \textbf{54}(1),  137--178 (2021)

\bibitem{tian2020prior}
Tian, Z., Zhao, H., Shu, M., Yang, Z., Li, R., Jia, J.: Prior guided feature
  enrichment network for few-shot segmentation. IEEE Transactions on Pattern
  Analysis and Machine Intelligence pp.~1--1 (2020)

\bibitem{triantafillou2017few}
Triantafillou, E., Zemel, R., Urtasun, R.: Few-shot learning through an
  information retrieval lens. In: Advances in Neural Information Processing
  Systems. pp. 2252--2262 (2017)

\bibitem{vaswani2017attention}
Vaswani, A., Shazeer, N., Parmar, N., Uszkoreit, J., Jones, L., Gomez, A.N.,
  Kaiser, {\L}., Polosukhin, I.: Attention is all you need. In: Advances in
  Neural Information Processing Systems. pp. 5998--6008 (2017)

\bibitem{velivckovic2018graph}
Veličković, P., Cucurull, G., Casanova, A., Romero, A., Liò, P., Bengio, Y.:
  Graph attention networks. In: International Conference on Learning
  Representations (2018), \url{https://openreview.net/forum?id=rJXMpikCZ}

\bibitem{vinyals2016matching}
Vinyals, O., Blundell, C., Lillicrap, T., Wierstra, D., et~al.: Matching
  networks for one shot learning. Advances in Neural Information Processing
  Systems  \textbf{29},  3630--3638 (2016)

\bibitem{wang2020few}
Wang, H., Zhang, X., Hu, Y., Yang, Y., Cao, X., Zhen, X.: Few-shot semantic
  segmentation with democratic attention networks. In: Proceedings of the
  European Conference on Computer Vision. pp. 730--746. Springer (2020)

\bibitem{wang2019panet}
Wang, K., Liew, J.H., Zou, Y., Zhou, D., Feng, J.: {PAN}et: Few-shot image
  semantic segmentation with prototype alignment. In: Proceedings of the
  IEEE/CVF International Conference on Computer Vision. pp. 9197--9206 (2019)

\bibitem{yang2020prototype}
Yang, B., Liu, C., Li, B., Jiao, J., Ye, Q.: Prototype mixture models for
  few-shot semantic segmentation. In: Proceedings of the European Conference on
  Computer Vision. pp. 763--778. Springer (2020)

\bibitem{zhang2021self}
Zhang, B., Xiao, J., Qin, T.: Self-guided and cross-guided learning for
  few-shot segmentation. In: Proceedings of the IEEE/CVF Conference on Computer
  Vision and Pattern Recognition. pp. 8312--8321 (2021)

\bibitem{zhang2019pyramid}
Zhang, C., Lin, G., Liu, F., Guo, J., Wu, Q., Yao, R.: Pyramid graph networks
  with connection attentions for region-based one-shot semantic segmentation.
  In: Proceedings of the IEEE/CVF International Conference on Computer Vision.
  pp. 9587--9595 (2019)

\bibitem{zhang2019canet}
Zhang, C., Lin, G., Liu, F., Yao, R., Shen, C.: {CAN}et: Class-agnostic
  segmentation networks with iterative refinement and attentive few-shot
  learning. In: Proceedings of the IEEE/CVF Conference on Computer Vision and
  Pattern Recognition. pp. 5217--5226 (2019)

\bibitem{zhang2021few}
Zhang, G., Kang, G., Wei, Y., Yang, Y.: Few-shot segmentation via
  cycle-consistent {T}ransformer. arXiv preprint arXiv:2106.02320  (2021)

\bibitem{zhang2020sg}
Zhang, X., Wei, Y., Yang, Y., Huang, T.S.: {SG-One}: Similarity guidance
  network for one-shot semantic segmentation. IEEE Transactions on Cybernetics
  \textbf{50}(9),  3855--3865 (2020)

\bibitem{zhang2021rethinking}
Zhang, Y., Mehta, S., Caspi, A.: Rethinking semantic segmentation evaluation
  for explainability and model selection. arXiv preprint arXiv:2101.08418
  (2021)

\bibitem{zhao2017pyramid}
Zhao, H., Shi, J., Qi, X., Wang, X., Jia, J.: Pyramid scene parsing network.
  In: Proceedings of the IEEE/CVF Conference on Computer Vision and Pattern
  Recognition. pp. 2881--2890 (2017)

\bibitem{zhu2021unified}
Zhu, F., Zhu, Y., Zhang, L., Wu, C., Fu, Y., Li, M.: A unified efficient
  pyramid {T}ransformer for semantic segmentation. In: Proceedings of the
  IEEE/CVF International Conference on Computer Vision. pp. 2667--2677 (2021)

\end{thebibliography}
\newpage
\begin{center}
\textbf{\large Supplementary Material:\\Dense Cross-Query-and-Support Attention Weighted Mask Aggregation for Few-Shot Segmentation}
\end{center}
\setcounter{equation}{0}
\setcounter{figure}{0}
\setcounter{table}{0}
\setcounter{page}{1}
\makeatletter
\renewcommand{\theequation}{S\arabic{equation}}
\renewcommand{\thefigure}{S\arabic{figure}}
\renewcommand{\thetable}{S\arabic{table}}
\renewcommand{\thepage}{S\arabic{page}}
{\color{purple}
\subsection*{A. More Results}
\textbf{Stability and robustness to trials and hyperparameter.}
In addition to the 4-fold cross validation in Table 1, we conduct three repeated trials on Fold-0 of PASCAL-5$^i$
and obtain stable IoUs (72.1$\pm$0.25).
We further experiment with
three hyperparamter pairs of batch size and learning rate (\textit{bs}/\textit{lr})
and obtain stable IoUs, too:
72.2 (48/0.001), 71.9 (24/0.001), and 71.9 (24/0.0005).

\textbf{Region-wise over- and under-segmentation measures.}
We further employ the region-wise over-segmentation measure (ROM) and region-wise under-segmentation measure (RUM) \cite{zhang2021rethinking} for quantitative evaluation.
ROM and RUM are two novel threshold-free metrics assessing region-based over- and under-segmentation, and expected to lend greater explainability to semantic segmentation performance in real-world applications.
A smaller ROM (RUM) value indicates less over- (under-) segmentation and is preferred.
The 1-shot ROM and RUM values are tabulated in Table~\ref{tab:ROM_RUM}, where our DCAMA slightly outperforms the competent HSNet in both metrics overall.

\begin{table}[h]
\caption{One-shot evaluation results in terms of ROM and RUM \cite{zhang2021rethinking}.
HSNet$^\dagger$: our reimplementation based on the official codes.}\label{tab:ROM_RUM}
\centering
\begin{tabular}{c|c|cccccccc|cc}
\hline
\multirow{2}{*}{Method} & \multirow{2}{*}{Backbone} & \multicolumn{2}{c}{PASCAL-5$^i$} &  & \multicolumn{2}{c}{COCO-20$^i$} &  & \multicolumn{2}{c|}{FSS-100} & \multicolumn{2}{c}{Overall} \\ \cline{3-4} \cline{6-7} \cline{9-12}
                        &                           & ROM             & RUM            &  & ROM            & RUM            &  & ROM           & RUM          & ROM          & RUM          \\ \hline
HSNet$^\dagger$~\cite{min2021hypercorrelation} & Swin-B                    & 0.26            & \textbf{0.06}           &  & 0.15           & 0.07           &  & \textbf{0.12}          & 0.03         & 0.177 & 0.053 \\
DCAMA                   & Swin-B                    & \textbf{0.21}            & 0.07           &  & \textbf{0.13}           & \textbf{0.06}           &  & 0.15          & \textbf{0.02}         & \textbf{0.163} & \textbf{0.050} \\ \hline
\end{tabular}
\end{table}

\textbf{Computational efficiency.}
In addition to what is reported in the main text, below we provide more metrics regarding computational efficiency in Table \ref{tab:efficiency}.
Our DCAMA has slightly larger FLOPS but comparable times with HSNet for training an epoch and inference, and converges in substantially fewer epochs thus needing much less time for training.

\begin{table}[h]
    \caption{Computational efficiency (1-shot on COCO-20$^i$).
    HSNet$^\dagger$: our reimplementation based on the official codes.}
    \label{tab:efficiency}
    \centering
    \begin{tabular}{cccccc}
    \hline
        Method & Backbone & Epoch to converge & Epoch time & FLOPS & Inference time \\ \hline
        HSNet$^\dagger$~\cite{min2021hypercorrelation}  & Swin-B   &355                & $\sim$4 min            &103.8 G       &0.13 s                \\
        DCAMA  & Swin-B   &90     & $\sim$4 min            &109.4 G       &0.13 s                \\ \hline
    \end{tabular}
\end{table}

{\bf Memory and latency analysis for $n$-shot inference.}
The analysis for $n=1$ to $5$ is presented in Fig. \ref{fig:latency}.
The increases in memory and latency are approximately linear with $n$.

\begin{figure}[t]
  \centering
  \includegraphics[trim=663 0 0 0, clip, width=.8\linewidth]{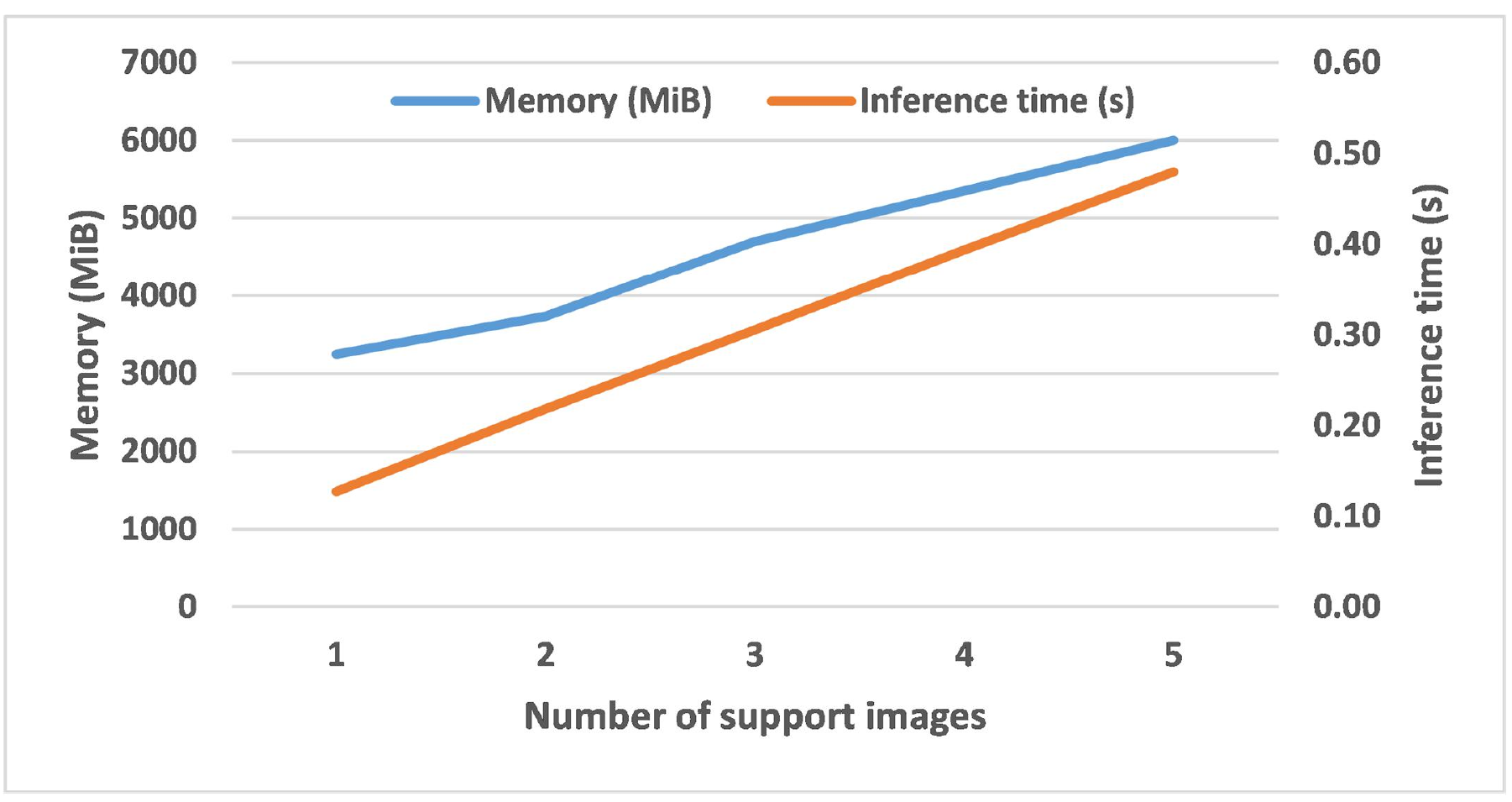}
   \caption{Memory and latency analysis for $n$-shot inference.}
   \label{fig:latency}
\end{figure}
}

\begin{table}[b]
\caption{Ablation study on feature skip-connection configuration (1-shot on PASCAL-5$^i$~\cite{shaban2017one} with Swin-B \cite{liu2021swin} as backbone).}\label{tab:skipconnect}
\centering
\begin{tabular}{ccc|cccc|c|c}
\hline
\multicolumn{3}{c|}{Feature scale}                                                                                             &                          &                          &                          &                          &                        &                          \\ \cline{1-3}
\multicolumn{1}{c|}{1/4}                                & \multicolumn{1}{c|}{1/8}                                & 1/16       & \multirow{-2}{*}{Fold-0} & \multirow{-2}{*}{Fold-1} & \multirow{-2}{*}{Fold-2} & \multirow{-2}{*}{Fold-3} & \multirow{-2}{*}{mIoU} & \multirow{-2}{*}{FB-IoU} \\ \hline
\multicolumn{1}{c|}{}                                   & \multicolumn{1}{c|}{}                                   &            & 70.6                     & 72.6                     & 61.4                     & 64.5                     & 66.2                   & 77.4                     \\
\multicolumn{1}{c|}{\checkmark}                         & \multicolumn{1}{c|}{}                                   &            & 71.0                     & 73.3                     & 61.7                     & 67.0                     & 68.2                   & 78.1                     \\
\multicolumn{1}{c|}{}                                   & \multicolumn{1}{c|}{\checkmark}                         &            & 70.7                     & \textbf{74.0}            & 63.9                     & 65.7                     & 68.6                   & 77.9                     \\
\rowcolor[HTML]{EFEFEF}
\multicolumn{1}{c|}{\cellcolor[HTML]{EFEFEF}\checkmark} & \multicolumn{1}{c|}{\cellcolor[HTML]{EFEFEF}\checkmark} &            & \textbf{72.2}            & 73.8                     & \textbf{64.3}            & \textbf{67.1}            & \textbf{69.3}          & \textbf{78.5}            \\
\multicolumn{1}{c|}{}                                   & \multicolumn{1}{c|}{}                                   & \checkmark & 66.7                     & 62.6                     & 49.6                     & 60.1                     & 59.8                   & 71.9                     \\
\multicolumn{1}{c|}{\checkmark}                         & \multicolumn{1}{c|}{\checkmark}                         & \checkmark & 67.3                     & 59.6                     & 50.6                     & 58.1                     & 58.9                   & 72.2                     \\ \hline
\end{tabular}
\end{table}

\subsection*{B. Further Ablation Studies}
\textbf{Configuration of skip connections.}
In Section 3.2 of the main text, we propose to skip connect (concatenate) extracted features of the input images to the integrated output of the multi-scale multi-layer Dense Cross-query-and-support Attention weighted Mask Aggregation (DCAMA) blocks, following successful experience of previous works~\cite{ronneberger2015u,zhao2017pyramid}.
Here, we empirically determine the optimal configuration of the skip connections, by comparing the effects of concatenating (or not) the $\frac{1}{4}$, $\frac{1}{8}$, and $\frac{1}{16}$ scale features individually and jointly.
As shown in Table~\ref{tab:skipconnect}, concatenating either $\frac{1}{4}$ or $\frac{1}{8}$ scale features individually brings notable performance improvement upon the baseline of no skip connection, and their joint concatenation brings further improvement to achieve the optimal performance ($+$3.1\% and $+$1.1\% with respect to the baseline in mIoU and FB-IoU, respectively).
On the other hand, concatenating the $\frac{1}{16}$ scale features, either individually or jointly with the shallower features, leads to obvious performance deterioration.
Similar findings were also reported in previous works~\cite{tian2020prior,zhang2019canet}, with the possible explanation that semantic information contained in high-level features is more class-specific and less generalizable.
Based on these results, we choose to skip connect the $\frac{1}{4}$ and $\frac{1}{8}$ scale features in our DCAMA framework for comparison with other methods.

\textbf{Contribution of multi-scale attention.}
In Section 3.2 of the main text, we implement the multi-layer DCAMA blocks at all scales (i.e., $\frac{1}{8}$, $\frac{1}{16}$, and $\frac{1}{32}$) allowed by our hardware, as the multi-scale strategy has proven effective in various computer vision applications.
Here, to empirically evaluate the impact of the multi-scale attention, we conduct experiments to ablate the multi-scale attention---one scale at a time, following \cite{min2021hypercorrelation}.
The results are shown in Table \ref{tab:multiscale}.
As we can see, removing $\frac{1}{8}$ scale attention results in modest and slight decreases in mIoU and FB-IoU by 2.1\% and 1.4\%, respectively, and consecutively removing $1/16$ scale attention leads to further, substantial performance degradation in both metrics by 8.5\% and 6.6\%, respectively.
These results indicate the indispensable role of the multi-scale attention to our proposed framework.

\begin{table}[h]
  \caption{Ablation study on the multi-scale attention strategy (1-shot on PASCAL-5$^i$~\cite{shaban2017one} with Swin-B \cite{liu2021swin} as backbone).}\label{tab:multiscale}
  \centering
\begin{tabular}{ccc|cccc|c|c}
\hline
\multicolumn{3}{c|}{Attention scale}                                                                                           &                          &                          &                          &                          &                        &                          \\ \cline{1-3}
\multicolumn{1}{c|}{1/8}                                & \multicolumn{1}{c|}{1/16}                               & 1/32       & \multirow{-2}{*}{Fold-0} & \multirow{-2}{*}{Fold-1} & \multirow{-2}{*}{Fold-2} & \multirow{-2}{*}{Fold-3} & \multirow{-2}{*}{mIoU} & \multirow{-2}{*}{FB-IoU} \\ \hline
\multicolumn{1}{c|}{}                                   & \multicolumn{1}{c|}{}                                   & \checkmark & 62.0                     & 66.8                     & 49.4                     & 56.7                     & 58.7                   & 70.5                     \\
\multicolumn{1}{c|}{}                                   & \multicolumn{1}{c|}{\checkmark}                         & \checkmark & 70.0                     & 73.3                     & 61.5                     & 64.0                     & 67.2                   & 77.1                     \\
\rowcolor[HTML]{EFEFEF}
\multicolumn{1}{c|}{\cellcolor[HTML]{EFEFEF}\checkmark} & \multicolumn{1}{c|}{\cellcolor[HTML]{EFEFEF}\checkmark} & \checkmark & \textbf{72.2}            & \textbf{73.8}            & \textbf{64.3}            & \textbf{67.1}            & \textbf{69.3}          & \textbf{78.5}            \\ \hline
\end{tabular}
\end{table}

{\color{purple}
\textbf{Impact of skip-connecting support features/the number of support foreground and background pixels.}
The ablation results in Table \ref{tab:support} show that both excluding support features from skip connection (row a) and normalizing foreground and background region sizes (to 600 pixels following CyCTR; row b) impair the performance.
Specifically, the high-level feature maps that are skip connected have relative large receptive fields, hence are helpful even not aligned.

\begin{table}[]
    \caption{Ablation study on skip-connecting support features and the number of support foreground and background pixels (1-shot on PASCAL-5$^i$~\cite{shaban2017one} with Swin-B \cite{liu2021swin} as backbone).}
    \label{tab:support}
    \centering
    \begin{tabular}{c|cc|cccc|c|c}
\hline
Ablation & Supp. feat. & Num pix. & Fold-0        & Fold-1        & Fold-2        & Fold-3        & mIoU          & FB-IoU        \\ \hline
a        & \ding{55}   & All      & 71.1          & 73.6          & 63.5          & \textbf{67.5} & 68.9          & 77.9          \\
b        & \checkmark  & 600      & 70.7          & 66.6          & 62.9          & 62.6          & 65.7          & 75.9          \\ \hline
\rowcolor[HTML]{EFEFEF}
DCAMA    & \checkmark  & All      & \textbf{72.2} & \textbf{73.8} & \textbf{64.3} & 67.1          & \textbf{69.3} & \textbf{78.5} \\ \hline
\end{tabular}
\end{table}
}

\subsection*{C. More Visual Analysis}
\begin{figure*}[!t]
  \centering
  \includegraphics[trim=30 0 0 0, clip, width=\textwidth]{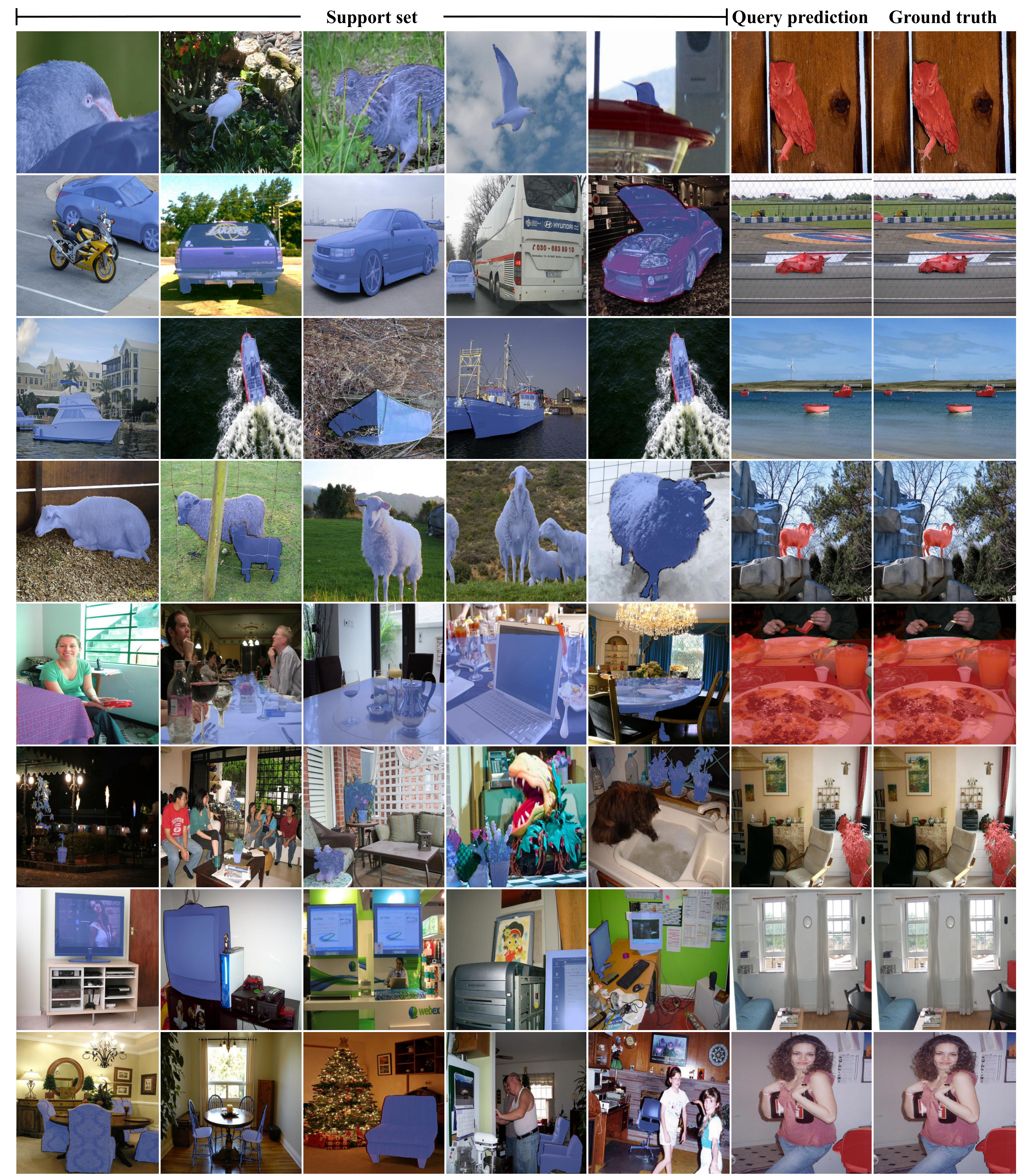}
  \caption{Example 5-shot segmentation results by the proposed DCAMA framework (with Swin-B \cite{liu2021swin} as backbone) on PASCAL-5$^i$, in the presence of intra-class variations, size differences, complex background, and occlusions.}
  \label{fig:5shotresults}
\end{figure*}

\textbf{Visualization of more segmentation results.}
Fig. \ref{fig:5shotresults} shows example 5-shot segmentation results by our proposed DCAMA framework, complementary to the 1-shot segmentation results shown in the main text.

\begin{figure*}[!t]
  \centering
  \includegraphics[trim=0 0 0 0, clip, width=\textwidth]{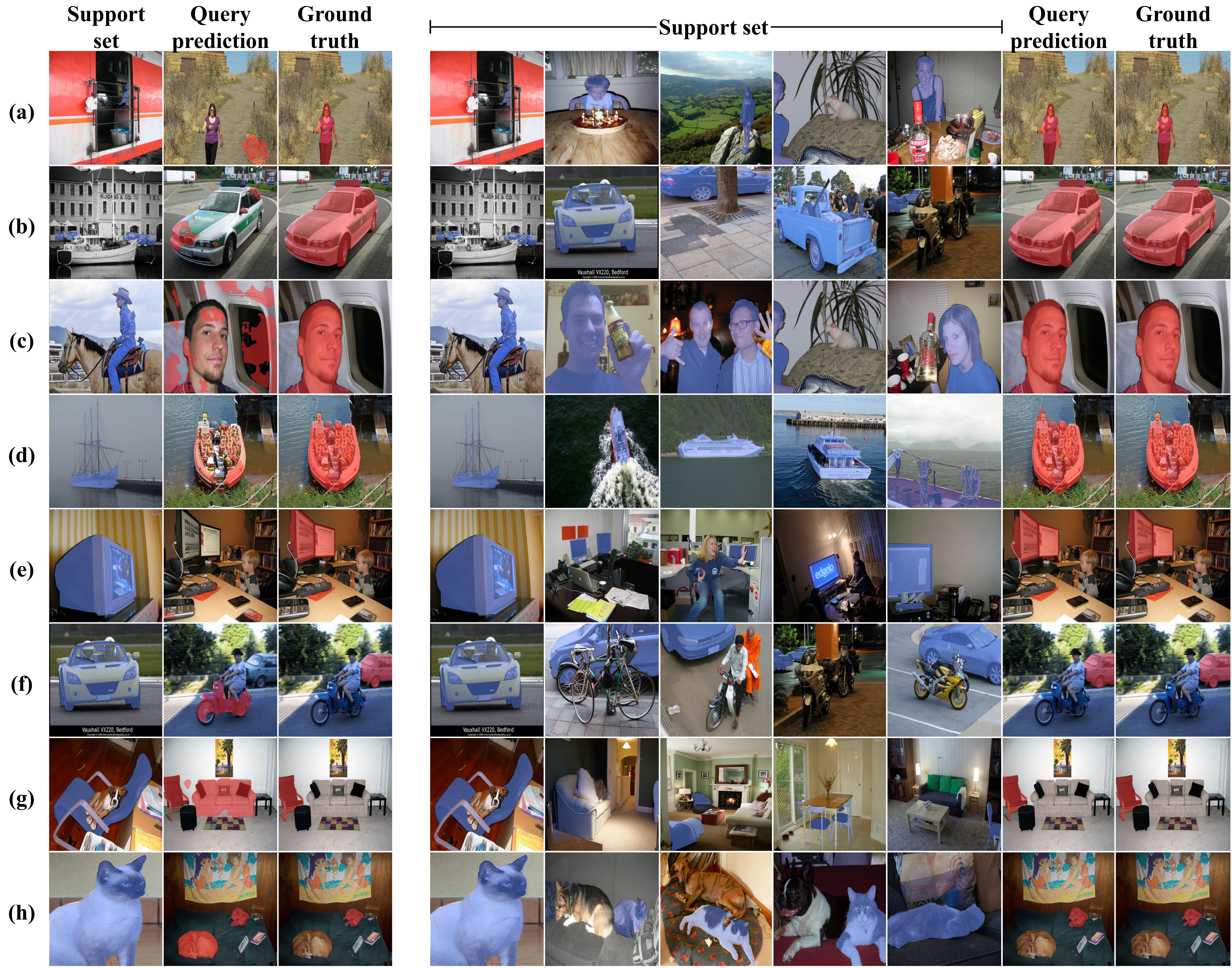}
  \caption{Left: some representative failure cases on PASCAL-5$^i$ in 1-shot setting.
  Right: the same cases in 5-shot setting, where the extra support images and masks help our DCAMA framework produce accurate segmentation of the query images in these challenging cases.}
  \label{fig:failure_5shotresults}
\end{figure*}

\textbf{Qualitative limitation analysis.}
We empirically explore the limitations of the proposed DCAMA framework, based on qualitative analysis of failure cases in the 1-shot settings.
Above all, as shown in Fig. \ref{fig:failure_5shotresults}, all the failure cases in the 1-shot setting are accurately segmented with the extra support images in the 5-shot setting, in accordance with the findings of Min et al.~\cite{min2021hypercorrelation}.
As this finding clearly demonstrates the efficacy of 5-shot segmentation and is expected, it is interesting to dig deeper to find out why the 1-shot segmentation fails in these cases and how the extra support images help.
Systematically, we roughly categorize the 1-shot failures into three types of limitations: limited representativeness of the support image, intra-class variation, and inter-class similarity, which sometimes occur together, too.

Limited representativeness happens when the target class is under-represented in the support image, e.g., the object is largely occluded (row (a) of Fig. \ref{fig:failure_5shotresults}), too small (row (b)), or in a quite different perspective (row (c)).
In contrast, for intra-class variation, although the object in the support image is complete and in normal size and view, the instance in the query image can look differently despite belonging to the same major class (rows (d) and (e)).
The third type of limitation is inter-class similarity, i.e., the similarity between different classes causing difficulty in differentiating them (rows (f)--(h)).
We can also observe its concurrence with the other two types (rows (f) and (h)).
These limitations are inherent in Few-Shot Learning (FSL) and faced by all FSL algorithms.
Theocratically, these limitations can be effectively overcome by introducing the missing support information with a few additional, informative support images.

\begin{figure*}[!t]
  \centering
  \includegraphics[trim=0 0 0 0, clip, width=\textwidth]{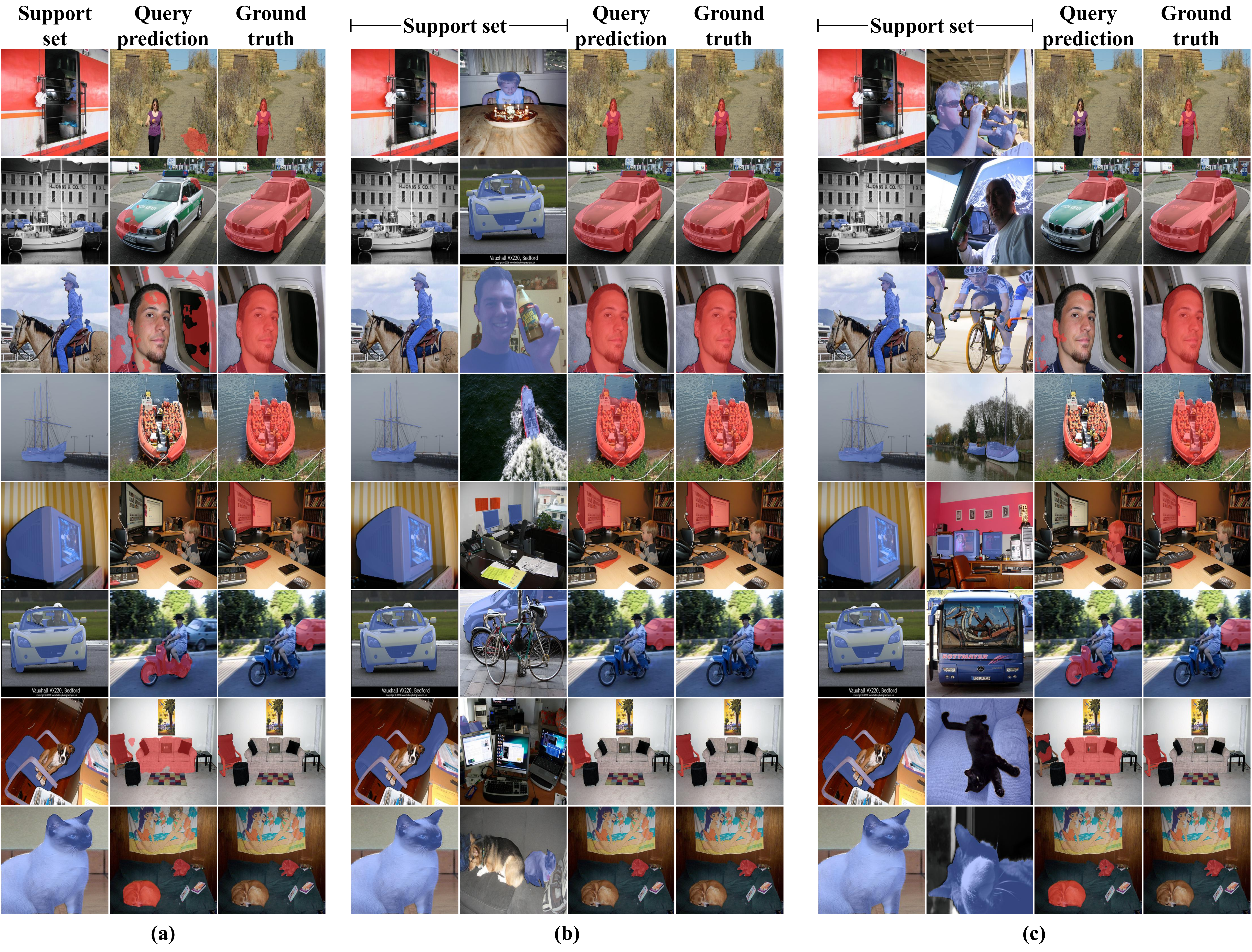}
  \caption{Control experiment on fixing 1-shot failures by adding one extra support image:
  (a) the same cases as shown in Fig. \ref{fig:failure_5shotresults} (left);
  (b) the failures are effectively fixed with an informative support image added; and
  (c) adding a non-informative support image helps little.}
  \label{fig:failure_fix}
\end{figure*}

To validate this, we pick only one extra informative support image and use it together with the original 1-shot support image for a 2-shot inference.
As shown in Fig. \ref{fig:failure_fix}(b), by providing more completed information about the target class and extra information about the within-class variance and inter-class differentiation, respectively, the 1-shot failures are fixed, as expected.
On the other hand, we also experiment with replacing the added informative support image with a non-informative one as a control group, and the corresponding results in Fig. \ref{fig:failure_fix}(c) are apparently inferior to those in Fig. \ref{fig:failure_fix}(b), with little or no improvement upon the original 1-shot results.
This suggests that the actual information contained in the support images may matter more than the absolute number of them.
To this end, it is desirable to integrate active learning with the few-shot segmentation for practical application.

\begin{figure}[t]
  \centering
  \includegraphics[width=.99\linewidth]{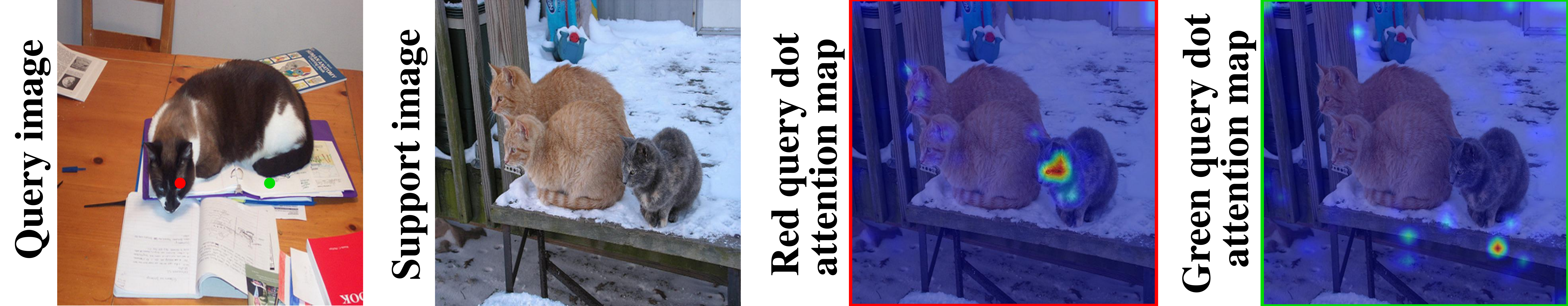}
  \caption{Point-wise attention weight maps for two pixels (red dot: foreground; and green dot: background) in the query image, to all pixels of a support image.
  The attention maps are obtained by averaging all the multi-scale multi-layer attention maps (upsampled where applicable).}\label{fig:attn}
\end{figure}

\textbf{Visualization of point-wise attention maps.}
For an intuitive perception of how a specific point in the query image correlates to all pixels of a support image, we visualize the attention weight maps for two query pixels---one foreground and one background---in Fig. \ref{fig:attn}.
As we can see, both of the query pixels have the strongest attention weights around their most similar regions in the support image, i.e., the dark cat's eye and below the dark cat's hind legs, respectively, while faint responses can also be observed around similar regions of the orange cats.

\begin{figure}[h]
  \centering
  \includegraphics[trim=35 0 598 0, clip, width=.8\linewidth]{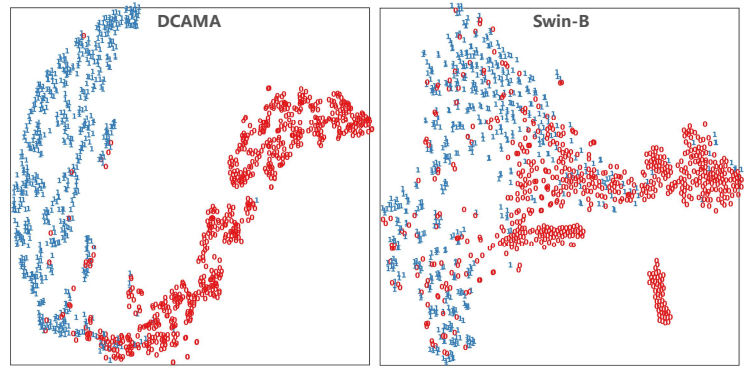}
   \caption{T-SNE plots for a testing class ``plane'': blue (red) indicates foreground (background).}
   \label{fig:t-SNE}
\end{figure}

{\color{purple}\textbf{T-SNE visualization.}
For an intuitive perception of how well the
model learns the metric space, we employ t-SNE for qualitative analysis (Fig. \ref{fig:t-SNE}).
The penultimate-layer features of DCAMA are more separable than those of the frozen backbone (Swin-B), indicating effective learning of the metric space.}

\end{document}